\documentclass[a4paper,fleqn]{cas-sc}
\usepackage[authoryear]{natbib}

\usepackage{hyperref}
\hypersetup{colorlinks,allcolors=black}
\usepackage{amssymb}
\usepackage{times}
\usepackage{url}
\usepackage[edges]{forest}
\usepackage{latexsym}
\usepackage{multirow}
\usepackage{array}
\usepackage{enumerate}
\usepackage{graphicx}
\usepackage{subfigure}
\usepackage{booktabs}
\usepackage{epstopdf}
\usepackage{amsmath}
\usepackage{color}

\usepackage{algorithmicx}
\usepackage{algpseudocode}
\usepackage{amsmath}
\usepackage{color,xcolor}
\usepackage{algorithm} 
\usepackage[normalem]{ulem}
\usepackage{bm}
\usepackage{amsopn}
\usepackage{caption}
\usepackage{pifont}
\usepackage{microtype}
\usepackage{amsfonts}
\usepackage{threeparttable}
\usepackage{multirow}
\usetikzlibrary{trees,positioning,shapes,shadows,arrows}
\tikzset{parent/.style={rounded corners=5pt,  align=center, fill=white, draw=red!50!black!90},
    child/.style={draw=red!50!black!90, align=center,text width=2.5cm,rounded corners=5pt},
    model/.style={align=center, fill=orange!30!yellow!10!, text width=8cm,rounded corners=5pt},
    chars/.style={align=center, text width=6cm,rounded corners=5pt}
}
\def\tsc#1{\csdef{#1}{\textsc{\lowercase{#1}}\xspace}}
\tsc{WGM}
\tsc{QE}
\tsc{EP}
\tsc{PMS}
\tsc{BEC}
\tsc{DE}

\begin{document}
\let\WriteBookmarks\relax
\def\floatpagepagefraction{1}
\def\textpagefraction{.001}

\address[1]{School of Informatics, Xiamen University,Xiamen,361005, Fujian,
China}
\address[2]{Institute of Artificial Intelligence, Xiamen University, 
Xiamen,361005, Fujian,China}
\address[3]{{Key Laboratory of Digital Protection and Intelligent Processing of Intangible Cultural Heritage of Fujian and Taiwan, Ministry of Culture and Tourism}, Xiamen,361005, {Fujian},{China}}
\address[4]{{Alibaba Group},{Hangzhou},{310023}, {Zhejiang},{China}}

\author[1,3]{Binbin Xie}[orcid=0000-0001-7832-1507, style=chinese]\cormark[1]
\ead{xdblb@stu.xmu.edu.cn}
\author[2,3]{Jia Song}[orcid=0000-0002-0415-7748, style=chinese]\cormark[1]
\ead{songjia@stu.xmu.edu.cn}
\author[1,3]{Liangying Shao}[orcid=0000-0002-2230-9952, style=chinese]\cormark[1]
\ead{liangyingshao@stu.xmu.edu.cn}
\author[3]{Suhang Wu}[style=chinese]
\ead{wush xmu@outlook.com}
\author[4]{Xiangpeng Wei}[style=chinese]
\ead{pemywei@gmail.com}
\author[4]{Baosong Yang}[style=chinese]
\ead{yangbaosong.ybs@alibaba-inc.com}
\author[4]{Huan Lin}[style=chinese]
\ead{lilai.lh@alibaba-inc.com}
\author[4]{Jun Xie}[style=chinese]
\ead{qingjing.xj@alibaba-inc.com}
\author[1,2,3]{Jinsong Su}[style=chinese, orcid=0000-0001-5606-7122]\cormark[2]
\ead{jssu@xmu.edu.cn}
\cortext[cor1]{Equally contribution.}
\cortext[cor2]{Corresponding authors. Tel.: +86 18750236638}

\shorttitle{From Statistical Methods to Deep Learning, Automatic Keyphrase Prediction: A Survey}
\shortauthors{Binbin Xie et~al.}
\title[mode=title]{From Statistical Methods to Deep Learning, Automatic Keyphrase Prediction: A Survey}\tnotemark[1]
\tnotetext[1]{This work was supported in part by National Natural Science Foundation of China under Grant 62276219, in part by Natural Science Foundation of Fujian Province of China under Grant 2020J06001, and in part by Youth Innovation Fund of Xiamen under Grant 3502Z20206059. (Correspongding author: Jinsong Su.)}


\begin{abstract}
Keyphrase prediction aims to generate phrases (keyphrases) that highly summarizes a given document. 
Recently, researchers have conducted in-depth studies on this task from various perspectives. 
In this paper, we comprehensively summarize representative studies from the perspectives of dominant models, datasets and evaluation metrics.
Our work analyzes up to 167 previous works, achieving greater coverage of this task than previous surveys.
Particularly, 
we focus highly on deep learning-based keyphrase prediction, 
which attracts increasing attention of this task in recent years. 
Afterwards, 
we conduct several groups of experiments to carefully compare representative models. 
To the best of our knowledge, 
our work is the first attempt to compare these models using the identical commonly-used datasets and evaluation metric, 
facilitating in-depth analyses of their disadvantages and advantages. 
Finally, 
we discuss the possible research directions of this task in the future.
\end{abstract}
\maketitle

\section{Introduction}
With the rapid development of the Internet and the explosion of information,
how to efficiently acquire information from tremendous text data becomes more and more important. 
To do this, several information compression tasks have been proposed, such as automatic summarization and automatic keyphrase prediction.
Compared with other tasks,
automatic keyphrase prediction 
brings forward a higher request to the ability of information compression,
since it aims to automatically produce a few keyphrases representing the core contents of the input document.
As keyphrases can facilitate understanding documents and provide useful information to downstream
tasks, such as 
information retrieval \citep{gutwin1999improving},
document classification \citep{hammouda2005corephrase, DBLP:conf/acl/HulthM06}, document summarization \citep{DBLP:journals/wias/ZhangZM04, DBLP:conf/acl/WangC13, DBLP:conf/naacl/PasunuruB18}, question generation \citep{subramanian-etal-2018-neural} and opinion mining \citep{wilson-etal-2005-recognizing, berend-2011-opinion}, automatic keyphrase prediction has attracted increasing attention.

Table \ref{table:present-cn-main-performance} shows an example of automatic keyphrase prediction. Generally, keyphrases can be divided into two categories: \emph{present keyphrases}
that continuously appear
in the input document and
\emph{absent keyphrases} that do not match any contiguous subsequence of the document.
To achieve high-quality keyphrase prediction, early studies mainly focus on \emph{automatic keyphrase extraction} \citep{hulth-2003-improved,DBLP:conf/emnlp/MihalceaT04,DBLP:conf/icadl/NguyenK07,DBLP:conf/aaai/WanX08}, which aims to directly extract keyphrases from the input document. 
Recently, the rise of deep learning prompts researchers to focus on \emph{automatic keyphrase generation}
\citep{meng-etal-2017-deep,yuan-etal-2020-one,DBLP:conf/acl/YeGL0Z20}, where dominant models can generate not only present but also absent keyphrases. 
Tables \ref{tab:papers} shows the number of papers related to automatic keyphrase prediction, published at 
the main computer science conferences.
It can be said that automatic keyphrase prediction has always been one of the research hotpots.

\begin{table}[t]
	\centering
	\renewcommand\arraystretch{1.5}
	\setlength{\tabcolsep}{1.5mm}{
        \caption{An example of keyphrase prediction and present keyphrases that appear in the document are underlined.}
		\begin{threeparttable}[width=0.95\columnwidth]
			\begin{tabular}{|p{0.95\columnwidth}|}
			\hline
			
				{\textbf{\emph{Input Document:}} A {nonmonotonic} observation {logic}. A variant of Reiter's {default logic} is proposed as a {logic} for reasoning with \underline{defeasible observations}. Traditionally, default rules are assumed to represent generic information and the facts are assumed to represent specific information about the situation, but in this paper, the specific information derives from \underline{defeasible observations} represented by (normal free) default rules, and the facts represent (hard) background knowledge. Whenever the evidence underlying some observation is more refined than the evidence underlying another observation, this is modelled by means of a priority between the default rules representing the observations. We thus arrive at an interpretation of prioritized normal free default logic as an observation logic, and we propose a semantics for this observation logic. Finally, we discuss how the proposed observation logic relates to the multiple extension problem and the problem of sensor fusion.}  \cr
			    \hline
			    {\textbf{\emph{Keyphrases:\  }}\underline{defeasible observations}; {{nonmonotonic logic;  prioritized default logic}} } \cr
				\hline
			\end{tabular}
		\end{threeparttable}
		\label{table:present-cn-main-performance}}
\end{table}

\begin{table}[htbp]
 \caption{Paper publications of keyphrase extraction and keyphrase generation at the main computer science conferences, and `--' denotes that the conference is not held or has not been held yet.}
  \centering
    \begin{tabular}{c|c|c|c|c|c|c}
    \toprule
    \textbf{Conf.} & \textbf{2017} & \textbf{2018} & \textbf{2019} & \textbf{2020} & \textbf{2021} & \textbf{2022} \\
    \midrule
    \multicolumn{7}{c}{Keyphrase Extraction} \\
    \midrule
    ACL   & 2     & 0     & 1     & 0     & 0     & 0 \\
    \midrule
    EMNLP & 0     & 0     & 1     & 1     & 3     & 0 \\
    \midrule
    NAACL & -- & 1     & 1     & -- & 2     & 1 \\
    \midrule
    COLING & -- & 0     & -- & 4     & -- & 3 \\
    \midrule
    AAAI  & 3     & 0     & 0     & 0     & 0     & 0 \\
    \midrule
    \multicolumn{7}{c}{Keyphrase Generation} \\
    \midrule
    ACL   & 1     & 0     & 3     & 3     & 2     & 0 \\
    \midrule
    EMNLP & 0     & 2     & 0     & 3     & 3     & 5 \\
    \midrule
    NAACL & -- & 0     & 2     & -- & 3     & 2 \\
    \midrule
    COLING & -- & 0     & -- & 1     & -- & 0 \\
    \midrule
    AAAI  & 0     & 0     & 1     & 0     & 1     & 2 \\
    \specialrule{0em}{1pt}{1pt}
    \hline\hline
    \specialrule{0em}{1pt}{1pt}
    \textbf{Total.} & 6     & 3     & 9     & 11    & 14    & 13 \\
    \bottomrule
    \end{tabular}%
  \label{tab:papers}%
\end{table}%

In this paper, 
we first provide a comprehensive review of automatic keyphrase prediction from the following aspects:
dominant models, datasets and evaluation metrics.
Compared with previous surveys \citep{DBLP:conf/acl/HasanN14,siddiqi2015keyword,ccano2019keyphrase,
alami2020automatic, nasar2019textual}, our work summarizes up to 167 previous works, achieving greater coverage of this task.
More importantly,
our work is not only the first attempt to thoroughly summarize keyphrase extraction based on neural networks, 
but also focusing highly on the recent advancements of neural keyphrase generation on different investigated problems.
Please note that neural keyphrase generation has become the hot research topic in this community, 
since it is able to predict not only present keyphrases but also absent keyphrases, 
which accounts a large proportion in the commonly-used keyphrase generation datasets. 
Particularly,
we further introduce the recent advancements in keyphrase generation, 
including pre-trained model based keyphrase generation models, 
echoing with the development trend of natural language processing.

Then, 
we conduct several groups of experiments to carefully compare representative models, so as to analyze their characteristics.
Unlike previous studies generally using different datasets and metrics to evaluate models, 
we use the identical commonly-used datasets and evaluation metric to ensure fair comparions among these representative models, 
and then analyze their advantages and disadvantages in different scenarios. 
Via our experiments, 
we can reach some interesting conclusions: 
1) Generally, unsupervised extraction models perform worst among all kinds of unsupervised and supervised models. However, when it exists a serious domain discrepancy between the training set and test set, the unsupervised extraction models may achieve comparable performance with the supervised ones. 
2) Among three commonly-used paradigms for keyphrase generation, ONE2SET surpasses the others and achieve the best performance, while is still inferior to the extraction models in predicting present keyphrases. 
3) Combining with extraction, generation and retrieval-based methods have potential to achieve better overall results for both present and absent keyphrase
predictions.

Finally, 
we point out the future research directions of keyphrase prediction task, which will play a positive role in guiding the follow-up studies. 
Note that we propose some directions that were not considered in previous surveys,
such as multi-modality keyphrase prediction, and multilingual keyphrase prediction.

\begin{figure*}[t]
\begin{minipage}[t]{1.0\textwidth}
\resizebox{\textwidth}{!}{
        \begin{forest}
        for tree={grow=east, forked edges,
            draw, rounded corners,
            node options={
                align=center },
            text width=2.7cm,
            anchor=west}
        [Automatic Keyphrase Extraction, parent
            [Hand-Engineered Features, for tree={child}
            [ Internal Document-based Features [{Statistical Features,
                    Position Features,
                    Linguistic Features,
                    Logical Structure},chars ]]
            [ External Document-Based Features  [{The Similarity based on Wiki, The Frequence based on External Documents, Citation,
            Web Linkage}, chars]]
            ]
            [ Models, for tree={child}
            [
            Unsupervised Models
                [Statistical Models, [\citep{DBLP:journals/is/El-BeltagyR09, DBLP:conf/ecir/0001MPJNJ18a, won2019automatic}, model]]
                [Graph-based Models, 
                    [Clustering, [\citep{DBLP:conf/adl/OhsawaBY98, grineva2009extracting, liu-etal-2009-clustering},model]]
                    [Graph Propagation, 
                        [Topic Information, [\citep{liu-etal-2010-automatic, DBLP:conf/www/SterckxDDD15, teneva-cheng-2017-salience, DBLP:conf/ijcnlp/BougouinBD13, boudin-2018-unsupervised}, model]]
                        [Others, [\citep{DBLP:conf/emnlp/MihalceaT04, DBLP:conf/aaai/WanX08, danesh-etal-2015-sgrank, florescu-caragea-2017-positionrank, DBLP:conf/aaai/GollapalliC14, DBLP:journals/ipm/Vega-OliverosGM19}, model]]
                    ]
                ]
                [Deep Learning-based Models,
                    [Phrase-Document Similarity, [\citep{DBLP:journals/ipm/Papagiannopoulou18, DBLP:journals/corr/abs-1801-04470, DBLP:journals/access/SunQZWZ20, DBLP:conf/aaai/LiD21}, model]]
                    [Graph-based Ranking, [\citep{DBLP:conf/naacl/MahataKSZ18, asl2020gleake, DBLP:journals/corr/abs-2111-07198, DBLP:conf/emnlp/LiangW0L21}, model]]
                    [Semantic Importance of Keyphrase, [\citep{DBLP:journals/corr/abs-2110-06651, joshi2022unsupervised}, model]]
                    [Attention Mechanism Information, [\citep{ding-luo-2021-attentionrank, gu2021ucphrase}, model]]]
            ]
            [
            Supervised Models
            [Statistical Models
                [Modeling Approaches [Sequence Labeling, [\citep{zhang2008automatic, DBLP:conf/aaai/GollapalliLY17}, model]]
                [Binary Classification, [\citep{DBLP:conf/dl/WittenPFGN99, DBLP:conf/ijcai/FrankPWGN99, DBLP:journals/corr/cs-LG-0212013, hulth-2003-improved,  DBLP:conf/ijcai/KelleherL05, DBLP:conf/www/YihGC06, medelyan2006thesaurus, DBLP:conf/waim/ZhangXTL06,  DBLP:conf/icadl/NguyenK07, shi2008improving, DBLP:conf/emnlp/MedelyanFW09, lopez-romary-2010-humb,  nguyen-luong-2010-wingnus,  DBLP:journals/jis/HaddoudA14, caragea-etal-2014-citation, DBLP:journals/kbs/XieWZ17, wang-li-2017-pku}, model]]
                [Ranking, [\citep{DBLP:conf/sigir/JiangHL09, DBLP:conf/cikm/ZhangCLG0X17}, model]]
                ]
                [ Algorithms,
                [ CRF, [\citep{zhang2008automatic, DBLP:conf/aaai/GollapalliLY17, 8844794}, model]] 
                [ Logistic Regression, [\citep{DBLP:conf/www/YihGC06, shi2008improving, DBLP:journals/jis/HaddoudA14}, model]] 
                [ Naive Bayes, [\citep{DBLP:conf/dl/WittenPFGN99, DBLP:conf/ijcai/FrankPWGN99, DBLP:conf/ijcai/KelleherL05, medelyan2006thesaurus, DBLP:conf/icadl/NguyenK07, nguyen-luong-2010-wingnus, caragea-etal-2014-citation, DBLP:journals/kbs/XieWZ17}, model]] 
                [ SVM, [\citep{DBLP:conf/waim/ZhangXTL06, DBLP:conf/sigir/JiangHL09}, model]] 
                 [ Bagged Decision Trees, [\citep{DBLP:journals/corr/cs-LG-0212013, DBLP:conf/emnlp/MedelyanFW09, lopez-romary-2010-humb}, model]] 
                  [ Other Ensemble Models,[\citep{hulth-2003-improved, wang-li-2017-pku}, model]] 
                ]
                ]
                [Deep Learning-based Models,
                    [Modeling Approaches
                    [Sequence Labeling, [\citep{DBLP:conf/emnlp/ZhangWGH16, DBLP:conf/naacl/ZhangLSZ18, DBLP:conf/bionlp/SaputraMW18,  DBLP:conf/acl/ZhangZ19a, DBLP:conf/www/ChowdhuryCC19, DBLP:journals/corr/abs-1910-08840,  DBLP:journals/corr/abs-2009-07119, DBLP:conf/jcdl/0001K020, DBLP:conf/emnlp/WangFR20, DBLP:conf/coling/SantoshSBD20,  DBLP:conf/bionlp/GeroH21, DBLP:journals/corr/abs-2106-04939}, model]]
                    [Binary Classification, [\citep{DBLP:conf/icmlc/WangPH05, DBLP:conf/emnlp/XiongHXCO19, DBLP:conf/naacl/PrasadK19}, model]]
                    [Ranking, [\citep{DBLP:journals/corr/abs-1004-3274, DBLP:journals/corr/abs-2002-05407, DBLP:conf/nlpcc/SunLXLB21, DBLP:conf/emnlp/SongJX21}, model]]]
                    [Data Utilization, [\citep{DBLP:conf/emnlp/LuanOH17, DBLP:conf/coling/LaiBKT20, DBLP:conf/aclnut/LeiHMZ21, kontoulis-etal-2021-keyphrase}, model]]
                    ]
                ]
                ]  
            ]    
        \end{forest}}
\caption{The Taxonomy of Representative Studies on Automatic Keyphrase Extraction.}
\label{fig:ke}
\end{minipage}
\end{figure*}
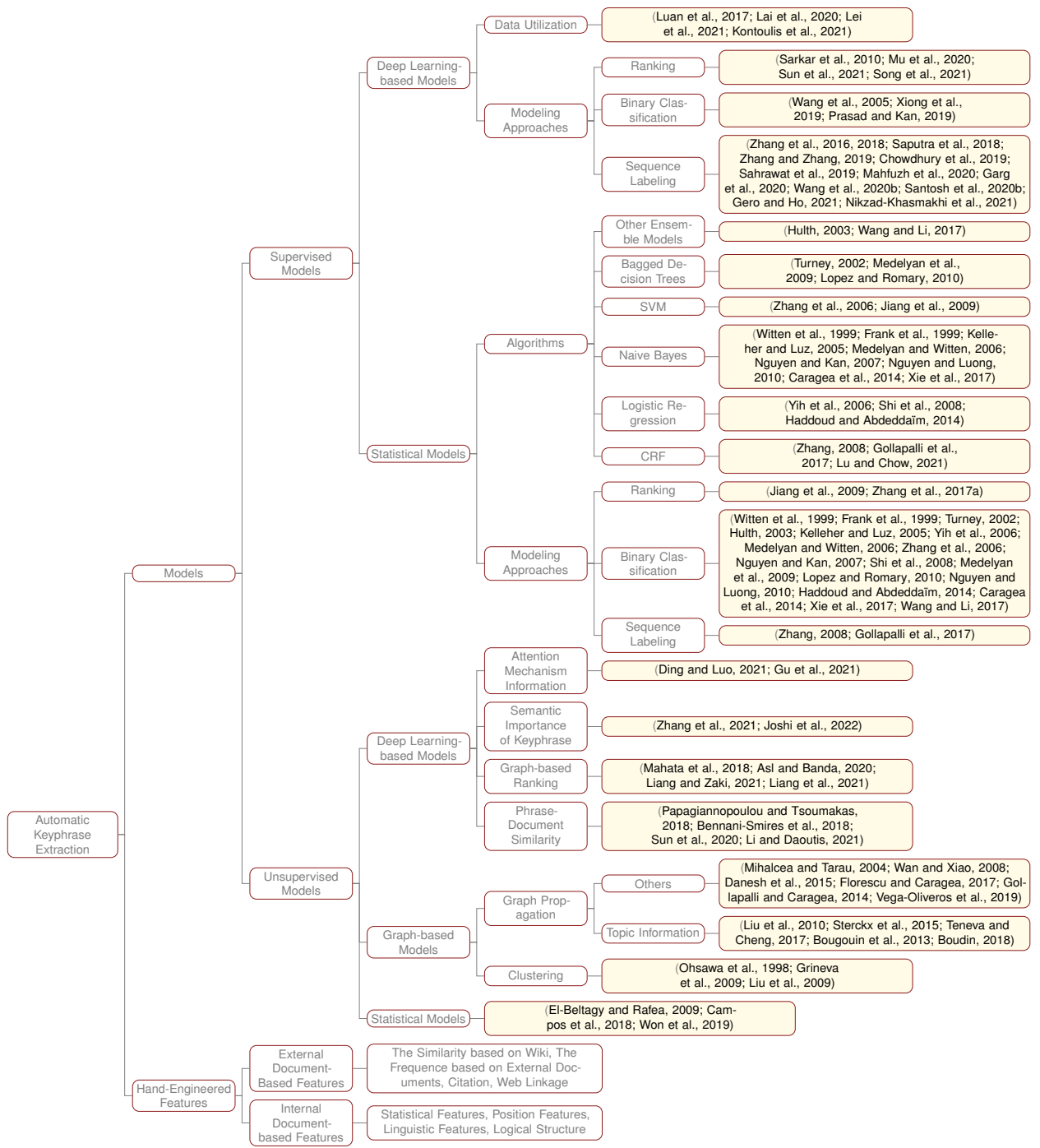

\section{Automatic Keyphrase Extraction}
Figure~\ref{fig:ke} shows the taxonomy of representative studies on automatic keyphrase extraction.
This line of research mainly focuses
on how to directly extract keyphrases from an input document.
Usually, it consists of three steps: 1) applying hand-crafted rules to obtain candidate phrases, 
such as removing stop words \citep{liu-etal-2009-clustering},
applying POS tagging \citep{DBLP:conf/emnlp/MihalceaT04},
extracting n-grams \citep{DBLP:conf/dl/WittenPFGN99}, and using knowledge bases \citep{DBLP:conf/acl/NguyenP09},
2) designing various hand-engineered features to represent candidate keyphrases,
and 3) determining the final keyphrases based on features using unsupervised or supervised models.

In the following subsections,
we will first
briefly introduce the
hand-engineered features,
and then describe the unsupervised and supervised models using these features in detail.

\subsection{Hand-engineered Features}

There are mainly four kinds of the internal document-based features used \citep{DBLP:conf/dl/WittenPFGN99, DBLP:journals/corr/cs-LG-0212013, hulth-2003-improved, DBLP:conf/waim/ZhangXTL06, DBLP:conf/ecir/0001MPJNJ18a, DBLP:conf/adl/OhsawaBY98}:
statistical features (phrase length, TF-IDF, the number of sentences containing phrases, co-occurrence frequency, etc.), positional features (occurrence positions, sentence boundaries, etc.), linguistics features (POS tags, case information, surrounding words, etc.), and logical structure features (the hierarchy, title, author list of the input document, etc.).

In addition, many features are proposed using the external documents, such as the similarity based on Wikipedia, candidate frequency based on the external documents, citation and web linkage.

\subsection{Unsupervised Keyphrase Extraction}
Generally, unsupervised models for keyphrase extraction can be roughly divided into statistical models, graph-based models and deep learning-based models, which will be briefly introduced below.

\subsubsection{Unsupervised Statistical Models}
These models are directly conducted based on the abundant hand-engineered features.
Among these features, the most important one is TF-IDF \citep{DBLP:journals/ipm/SaltonB88},
which can quantify the importance of each candidate phrase and thus becomes the basis of many follow-up models.
For example, \cite{DBLP:journals/is/El-BeltagyR09} consider the position of each candidate in the input document and introduce a length-related weight to
adjust its TF-IDF value.
Furthermore, \cite{DBLP:conf/ecir/0001MPJNJ18a} propose YAKE involving five hand-engineered features: case information, phrase position, term frequency, the frequency of phrase appearing within different sentences, and the number of surrounding words.
Based on these features, \cite{won2019automatic} further determine the number of keyphrases according to the length of the input document.

\subsubsection{Unsupervised Graph-based Models}
KeyGraph \citep{DBLP:conf/adl/OhsawaBY98} is the first graph-based model for keyphrase extraction.
In this model, frequently co-occurrent phrases are connected to form a graph,
which is then partitioned into subgraphs via clustering.
Finally,
the importance of 
each candidate phrase is quantified according to the subgraph based statistical information.
\cite{grineva2009extracting} firstly calculate 
edge weights as the phrase-level semantic relatedness based on Wikipedia,
and then
apply the community detection algorithm \citep{NewmanGirvan2004} to obtain dense subgraphs, 
where phrases from the most important subgraphs are considered as keyphrases.
Similarly, \cite{liu-etal-2009-clustering}
construct a word graph and cluster words
according to the semantic distances based on the word co-occurrence frequency or Wikipedia statistics.
Then, the noun phrases expanded from cluster centers are chosen as keyphrases.

Inspired by PageRank \citep{page1999pagerank}, \cite{DBLP:conf/emnlp/MihalceaT04} propose TextRank that iteratively conducts importance propagation on a
co-occurrent word graph.
Along this line,
\cite{danesh-etal-2015-sgrank} extend TextRank by using phrases as graph nodes. Then, many features are explored to adjust edge weights, including statistical features (phrase frequency and length
\citep{danesh-etal-2015-sgrank}, word co-occurrence frequency \citep{DBLP:conf/aaai/WanX08}) and position information \citep{florescu-caragea-2017-positionrank}.
Besides, to exploit more contexts,
\cite{DBLP:conf/aaai/WanX08}, \cite{DBLP:conf/aaai/GollapalliC14} extend the single-document word graph with similar documents and citation network, respectively.
In addition to the PageRank-based centrality measure, \cite{DBLP:journals/ipm/Vega-OliverosGM19} consider other commonly-used centrality measures,and then propose an optimal combination of centrality measures to extract keywords from an undirected and unweighted word graph.


Intuitively,
ideal keyphrases should be consistent with the topics of the input document.
Thus,
researchers introduce
the topic information
to refine graph-based models.
Typically, \cite{liu-etal-2010-automatic} propose TPR that adopts LDA \citep{blei2003latent} to obtain topic information and then separately performs PageRank for each topic.
To alleviate the huge computational cost of TPR, researchers extend TPR into
Single Topical PageRank \citep{DBLP:conf/www/SterckxDDD15} and SalienceRank \citep{teneva-cheng-2017-salience}, 
both of which perform PageRank once for each document.
Compared to the former, the latter can extract not only topic-specific but also corpus-correlated keyphrases.
Unlike the above studies based on LDA, \cite{DBLP:conf/ijcnlp/BougouinBD13} propose TopicRank, which firstly clusters similar phrases to form topics and
then constructs a topic graph for PageRank.
Afterwards,
they select the most representative phrases from each topic as keyphrases.
To refine TopicRank,
\cite{boudin-2018-unsupervised} represents candidate phrases and topics in a single graph and
exploits their mutual reinforcement
to improve candidate ranking.

\subsubsection{Unsupervised Deep Learning-based Models}
With the prosperous development of deep learning, researchers introduce
neural networks 
to learn semantic representations of input documents and candidate phrases for ranking, of which studies can be roughly divided into the following four categories:
\textbf{Phrase-Document Similarity.} The common practice is to measure the importance of each candidate phrase according to the phrase-document representation similarity.
To do this,
EmbedRank \citep{DBLP:journals/corr/abs-1801-04470} uses Sent2Vec \citep{pagliardini-etal-2018-unsupervised} and Doc2Vec \citep{le2014distributed} to represent
candidates and input documents
as vectors.
As an extension, $\text{EmbedRank}^+$ additionally considers the similarities between 
candidates to generate diverse keyphrases.
Unlike EmbedRank using Sent2vec and Doc2vec,
SIFRank \citep{DBLP:journals/access/SunQZWZ20} defines the vector representations of
candidates,
sentences and
input documents
as weighted averages of their corresponding ELMo embeddings \citep{DBLP:conf/naacl/PetersNIGCLZ18},
respectively.
Further, $\text{SIFRank}^+$ considers the positions of candidates within the document.
Subsequently,
\cite{DBLP:conf/aaai/LiD21} improve SIFRank by incorporating domain relevance and phrase quality into ranking scores.
\cite{DBLP:journals/ipm/Papagiannopoulou18} 
use entire documents to learn Glove \citep{pennington2014glove} embeddings,
and then rank candidates according to the sum of word-document similarities.

\textbf{Graph-based Ranking.} Besides, researchers apply deep learning to
refine the unsupervised models based on phrase graphs.
For example,
Key2Vec \citep{DBLP:conf/naacl/MahataKSZ18}
directly trains FastText to
learn representations of candidate phrases and document themes,
and then uses candidate-theme similarities to adjust the edge weights of PageRank.
Similarly,
\cite{DBLP:journals/corr/abs-2111-07198} consider the co-occurrence and similarities between 
candidates
for more accurate edge weighting of PageRank.
Using embedding-based graph,
\cite{asl2020gleake} apply PageRank or centrality algorithm to obtain the importance of candidates for ranking.
\cite{DBLP:conf/emnlp/LiangW0L21} find that
the phrase-document representation similarity (i.e. EmbedRank) is insufficient to capture different contexts for
keyphrase extraction.
To address this issue, they
define a boundary-aware centrality to capture local salient information and positional information of candidates for ranking.

\textbf{Semantic Importance of Keyphrases.}
Keyphrases play an important role in the representation learning of the input document.
Thus,
the representation of the input document will change
if any keyphrase is missing.
To model this intuition,
\cite{DBLP:journals/corr/abs-2110-06651} alternatively mask each candidate phrase and evaluate its importance according to the representation difference between the original document and the masked one.
Recently, \cite{joshi2022unsupervised} adopt a similar strategy that mainly focuses on the change of topic distributions.

\textbf{Attention Mechanism Information.}
Different from the above studies based on deep learning similarities, 
\citep{ding-luo-2021-attentionrank} use self-attention weights to quantify the importance of each candidate phrase within the sentence
and measure its semantic relatedness to the document according to its cross-attention weights.
Additionally, \cite{gu2021ucphrase} generate pseudo keyphrases for unlabeled
documents
using unsupervised statistic models or an existing knowledge base, and then
train a
keyphrase classifier fed with the self-attention map from RoBERTa \citep{zhuang-etal-2021-robustly}.

\subsection{Supervised Keyphrase Extraction}
Usually,
supervised models for keyphrase extraction can be divided into statistical
and deep learning-based models.

\subsubsection{Supervised Statistical Models}
Similar to unsupervised
keyphrase extraction,
abundant supervised statistical models leverage well-designed features,
including
statistical features \citep{DBLP:conf/dl/WittenPFGN99, DBLP:journals/corr/cs-LG-0212013, DBLP:conf/ijcai/KelleherL05, DBLP:journals/jis/HaddoudA14, DBLP:journals/kbs/XieWZ17},
positional features \citep{DBLP:conf/ijcai/FrankPWGN99, medelyan2006thesaurus, zhang2008automatic, DBLP:conf/sigir/JiangHL09}, linguistic features \citep{hulth-2003-improved, DBLP:conf/aaai/GollapalliLY17}, logical structures \citep{DBLP:conf/www/YihGC06, DBLP:conf/waim/ZhangXTL06, DBLP:conf/icadl/NguyenK07, nguyen-luong-2010-wingnus}, and external document-based features
\citep{shi2008improving, DBLP:conf/emnlp/MedelyanFW09,lopez-romary-2010-humb,  DBLP:conf/aaai/GollapalliC14, wang-li-2017-pku, DBLP:conf/cikm/ZhangCLG0X17}.



Based on these features,
researchers model keyphrase extraction as a sequence labeling task \citep{zhang2008automatic, DBLP:conf/aaai/GollapalliLY17}, a binary classification task or a ranking task \citep{DBLP:conf/sigir/JiangHL09, DBLP:conf/cikm/ZhangCLG0X17}
with various machine learning algorithms, 
such as conditional random field \citep{zhang2008automatic, DBLP:conf/aaai/GollapalliLY17}, logistic regression \citep{DBLP:conf/www/YihGC06, shi2008improving, DBLP:journals/jis/HaddoudA14}, Naive Bayes \citep{DBLP:conf/dl/WittenPFGN99, DBLP:conf/ijcai/FrankPWGN99, DBLP:conf/ijcai/KelleherL05, medelyan2006thesaurus, DBLP:conf/icadl/NguyenK07, nguyen-luong-2010-wingnus, caragea-etal-2014-citation, DBLP:journals/kbs/XieWZ17}, SVM \citep{DBLP:conf/waim/ZhangXTL06}, bagged decision trees \citep{DBLP:journals/corr/cs-LG-0212013, DBLP:conf/emnlp/MedelyanFW09, lopez-romary-2010-humb} and other ensemble models \citep{hulth-2003-improved, wang-li-2017-pku}.

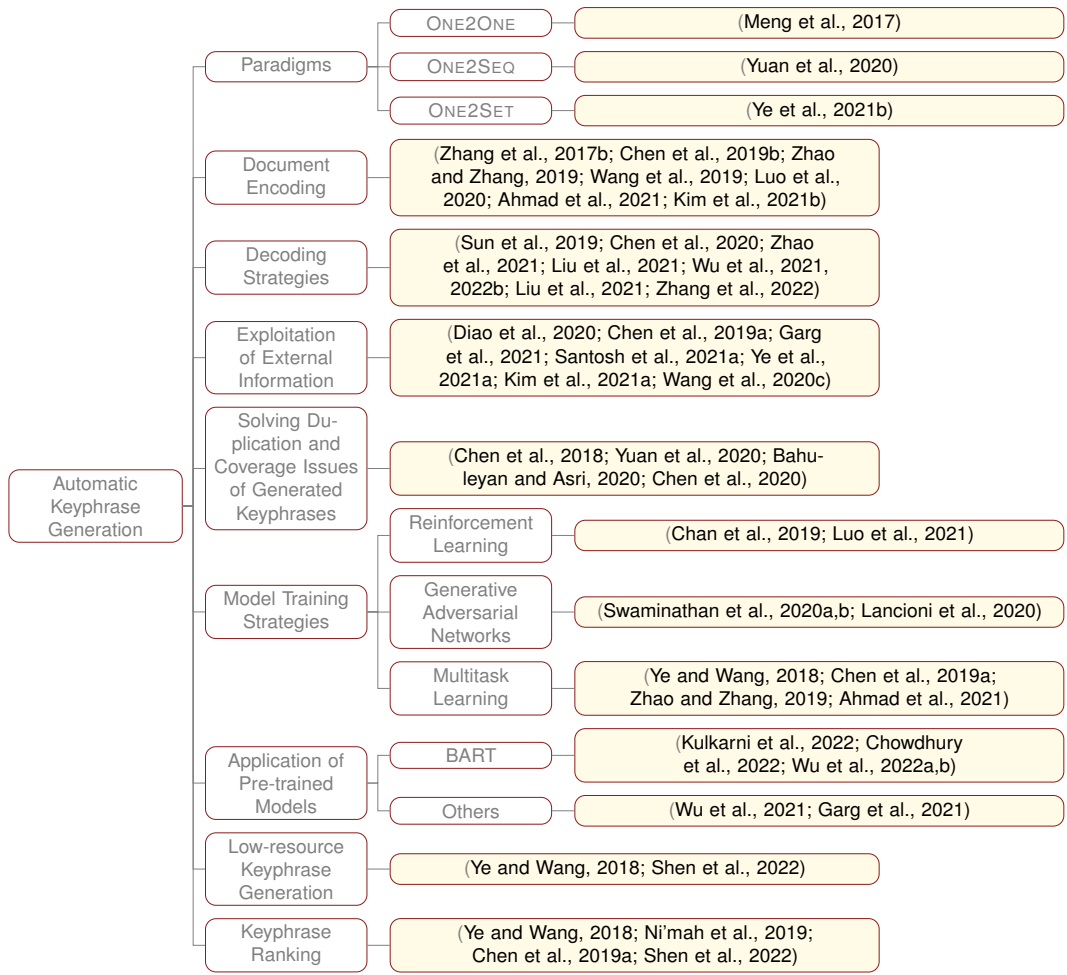
\begin{figure*}[t] \center
\begin{minipage}[t]{0.85\textwidth}
    \resizebox{\textwidth}{!}{%
        \begin{forest}
            for tree={
                grow'=east,
                forked edges,
                draw,
                rounded corners,
                node options={
                    align=center },
                text width=2.7cm,
                anchor=west,
            }
            [Automatic Keyphrase Generation, parent
                [
                Paradigms, for tree={child}
                    [\textsc{One2One}
                    [\citep{meng-etal-2017-deep}, model]]
                    [\textsc{One2Seq}[\citep{yuan-etal-2020-one}, model]]
                    [\textsc{One2Set}[\citep{DBLP:conf/acl/YeGL0Z20}, model]]
                ]
                [Document Encoding, for tree={child}[\citep{DBLP:conf/icsai/ZhangFX17, DBLP:conf/aaai/ChenGZKL19, DBLP:conf/acl/ZhaoZ19, wang-etal-2019-topic-aware,  DBLP:journals/corr/abs-2012-06754, ahmad-etal-2021-select, DBLP:conf/sigir/KimSH21}, model]]
                [Decoding Strategies, for tree={child}[\citep{DBLP:conf/sigir/SunTDDN19, DBLP:conf/acl/ChenCLK20, DBLP:conf/naacl/ZhaoBWWHZ21, DBLP:journals/taslp/LiuLW21, DBLP:conf/acl/WuLLNCZW21, wu2022fast, DBLP:journals/taslp/LiuLW21, DBLP:conf/sigir/ZhangJY0W22}, model]]
                [Exploitation of External Information, for tree={child}
                [\citep{DBLP:journals/corr/abs-2004-09800, DBLP:conf/naacl/ChenCLBK19,  DBLP:journals/corr/abs-2112-06776, DBLP:conf/pakdd/SantoshSBD21, DBLP:conf/emnlp/YeCGZ21, DBLP:conf/emnlp/KimJCH21,wang-etal-2020-cross-media}, model]]
                [Solving Duplication and Coverage Issues of Generated Keyphrases, for tree={child}[\citep{DBLP:conf/emnlp/ChenZ0YL18, yuan-etal-2020-one, DBLP:conf/coling/BahuleyanA20, DBLP:conf/acl/ChenCLK20}, model]]
                [Model Training Strategies, for tree={child}
                [Reinforcement Learning, [\citep{DBLP:conf/acl/ChanCWK19, DBLP:conf/emnlp/Luo0YQZ21}, model]
                ]
                [Generative Adversarial Networks, [\citep{DBLP:conf/aaai/SwaminathanGZMG20, swaminathan-etal-2020-preliminary, lancioni-etal-2020-keyphrase}, model]]
                [Multitask Learning, [\citep{DBLP:conf/emnlp/YeW18, DBLP:conf/naacl/ChenCLBK19, DBLP:conf/acl/ZhaoZ19,  ahmad-etal-2021-select}, model]]
                ]
                [Application of Pre-trained Models, for tree={child}
                [BART, [\citep{DBLP:journals/corr/abs-2112-08547, DBLP:journals/corr/abs-2201-05302, DBLP:journals/corr/abs-2203-08118, wu2022fast}, model]]
                [Others, [\citep{DBLP:conf/acl/WuLLNCZW21, DBLP:journals/corr/abs-2112-06776}, model]]
                ]
                [Low-resource Keyphrase Generation, for tree={child}[\citep{DBLP:conf/emnlp/YeW18, DBLP:journals/corr/abs-2104-08729}, model]]
                [Keyphrase Ranking, for tree={child}[\citep{DBLP:conf/emnlp/YeW18, DBLP:journals/corr/abs-1909-09485, DBLP:conf/naacl/ChenCLBK19,  DBLP:journals/corr/abs-2104-08729}, model]]
                ]     
            ]
        \end{forest}    
}
\caption{The Taxonomy of Representative Studies on Automatic Keyphrase Generation.}
\label{fig:kg}
\end{minipage}
\end{figure*}

\subsubsection{Supervised Deep Learning-based Models}
\cite{DBLP:conf/icmlc/WangPH05} first propose a feedforward neural network based classifier for supervised keyphrase extraction.
Henceforth,
deep learning-based supervised keyphrase extraction has gradually become one of the hot topics.

\textbf{Sequence Labeling.} 
Supervised keyphrase extraction is often modeled as a deep learning-based sequence labeling task.
Typically,
\cite{DBLP:conf/emnlp/ZhangWGH16} 
propose Joint-Layer RNN to extract keyphrases at different discrimination levels: 
judging whether the current word is a keyword and employing BIOES tagging scheme to identify keyphrases.
Based on Joint-Layer RNN,
\cite{DBLP:conf/naacl/ZhangLSZ18} introduce conversation context to
enrich the vector representations of microblog posts.
To simulate the human attention of reading during keyphrase annotating,
\cite{DBLP:conf/acl/ZhangZ19a} integrate an attention mechanism into Joint-Layer RNN.
Meanwhile, researchers also
explore more features for 
this model, such as medical concepts from an external knowledge base \citep{DBLP:conf/bionlp/SaputraMW18}, phonetics, phonological features \citep{DBLP:conf/www/ChowdhuryCC19}, and syntactical features \citep{DBLP:journals/corr/abs-2009-07119}.

Also, applying pre-trained models to supervised keyphrase extraction has become dominant.
For example,
on the basis of SciBERT \citep{DBLP:conf/emnlp/BeltagyLC19},
\citep{DBLP:journals/corr/abs-1910-08840} and \citep{DBLP:conf/jcdl/0001K020} stack BiLSTM+CRF and LSTM+CRF to
identify keyphrases, respectively.
Using the same model, \citep{DBLP:conf/ecir/SantoshSBD20} introduce a document-level attention and a gating mechanism to refine representation learning.
\cite{DBLP:conf/emnlp/WangFR20} separately leverage BERT and Transformer to encode the document and multi-modal information in web pages for keyphrase extraction.
\cite{DBLP:conf/bionlp/GeroH21}
use BERT-LSTM or BioBERT-LSTM to obtain the topic representations of input documents, encouraging the extraction of topic-consistent words.

Different from these studies, \cite{DBLP:conf/coling/SantoshSBD20} utilize graph encoders to separately incorporate syntactic and semantic dependency information for better encoder representation.
On the basis of the input document and the co-occurence graph,
\cite{DBLP:journals/corr/abs-2106-04939} adopt BERT and graph embedding techniques to learn the word-level textual and structure representations,
which are combined and fed into a sequence labeling tagger.


\textbf{Binary Classification.}
Researchers also
explore supervised keyphrase extraction as a binary classification task.
\cite{DBLP:conf/emnlp/XiongHXCO19} integrate the visual representation of the input document into ELMo word embeddings, and then use a convolutional Transformer to model interactions among candidate phrases for keyphrase classification. Besides, they introduce query prediction as a pre-training task.
\cite{DBLP:conf/naacl/PrasadK19} propose Glocal, an improved GCN, which incorporates the global importance of each node relative to other nodes
to learn word representations from a word graph.
Based on these representations,
keywords are identified via classification and finally used to reconstruct keyphrases via re-ranking. 

\textbf{Ranking.}
\cite{DBLP:journals/corr/abs-1004-3274} first apply a
deep learning-based ranking
model to achieve supervised keyphrase extraction.
\cite{DBLP:journals/corr/abs-2002-05407} use BERT stacked with BiLSTM to model semantic interactions among candidate phrases,
and then rank them according to the binary classification score and the hinge loss between the considered phrase and others.
\cite{DBLP:conf/nlpcc/SunLXLB21} propose JointKPE that learns to rank candidate phrases
according to
their document-level informativeness.
Particularly,
it is jointly trained with keyphrase chunking to guarantee the phraseness of candidates.
\cite{DBLP:conf/emnlp/SongJX21}
investigate three kinds of features for ranking: the syntactic accuracy of
the candidate phrase, the information saliency between the candidate and input document,
and the concept consistency between the candidate and the input document.

\textbf{Data Utilization.}
Based on a word graph,
\cite{DBLP:conf/emnlp/LuanOH17} employ label propagation together with a data selection scheme to leverage unlabeled documents.
\cite{DBLP:conf/coling/LaiBKT20} propose a self-distillation model
for keyphrase extraction.
In this approach,
a teacher model is trained on labeled examples,
while a student model is trained on both labeled examples and 
pseudo examples generated by the teacher model. 
During the subsequent training procedure,
the teacher model is re-initialized with the student model and repeats the above procedure.
To address the issue of incomplete annotated training data,
\cite{DBLP:conf/aclnut/LeiHMZ21} introduce negative sampling to adjust the training loss on unlabeled data.
From a different perspective,
\cite{kontoulis-etal-2021-keyphrase} 
believe that full-texts can provide richer information
while containing more
noise than the input abstract.
Thus, they leverage summaries induced from full-texts to refine keyphrase extraction.

\section{Automatic Keyphrase Generation}
Unlike the studies on keyphrase extraction,
keyphrase generation models can produce absent keyphrases that do not appear in the input document.
In this respect, \cite{meng-etal-2017-deep} propose the first keyphrase generation model, CopyRNN,
which inspires many subsequent models.
Usually,
these models are based on an encoder-decoder framework, where the encoder learns the semantic representation of each input document,
and then the decoder equipped with a copying mechanism \citep{gu-etal-2016-incorporating} automatically produces keyphrases.

In the following subsections,
we 
summarize representative advancements of keyphrase generation according to different investigated problems.
The taxonomy of representative studies on automatic keyphrase generation is shown in Figure~\ref{fig:kg}.

\begin{figure*}[t]
\centering
\subfigure[\textsc{One2One}]{
\includegraphics[width=1\linewidth]{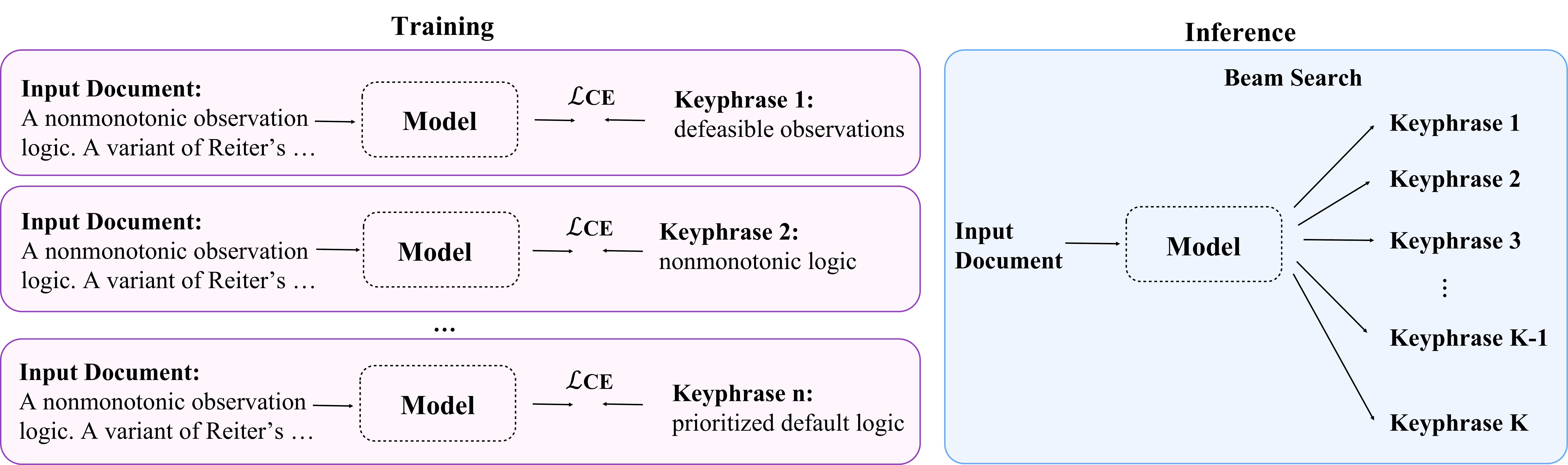}
}
\subfigure[\textsc{One2Seq}]{
\includegraphics[width=1\linewidth]{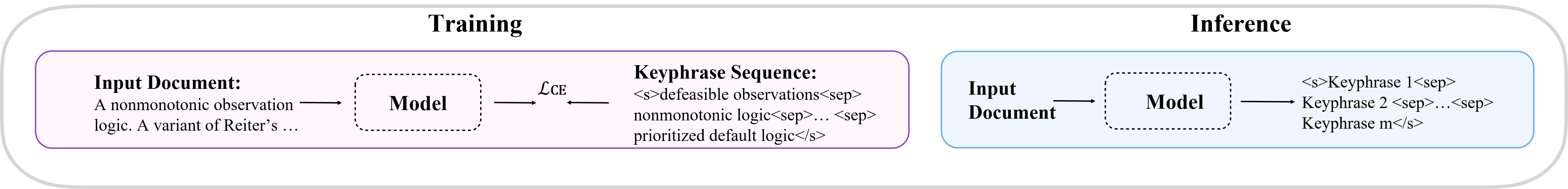}
}
\subfigure[\textsc{One2Set}]{
\includegraphics[width=1\linewidth]{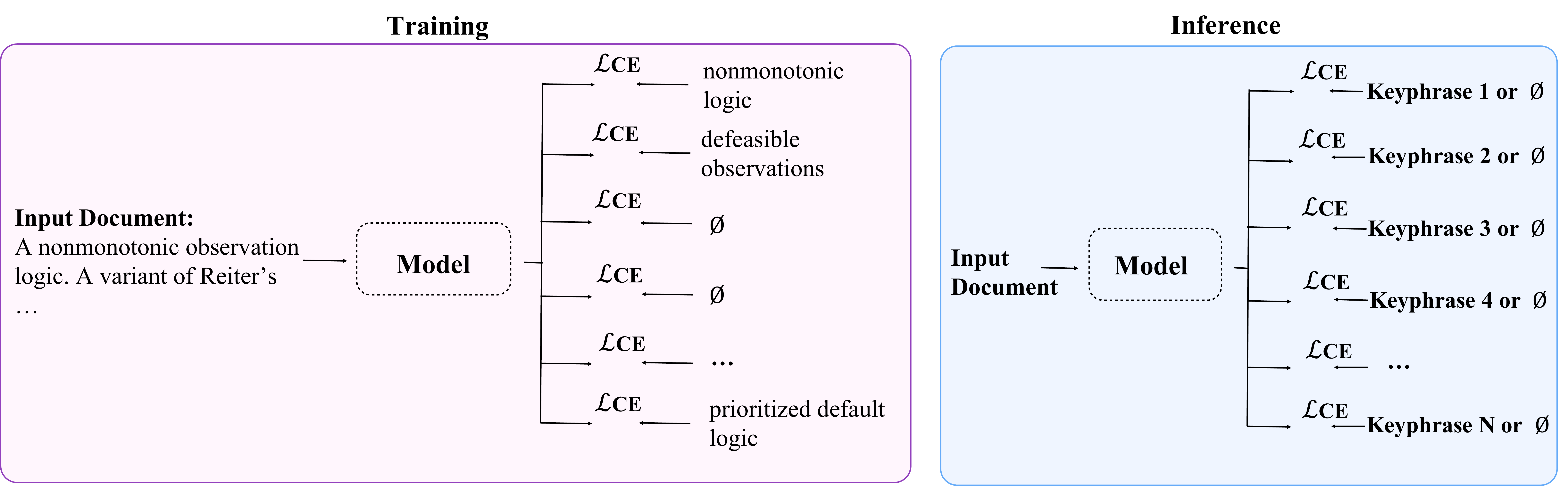}
}
\DeclareGraphicsExtensions.
\caption{The three dominant paradigms for keyphrase generation.}
\label{fig:paradigm}
\end{figure*}

\subsection{Paradigms}
Generally, paradigms of dominant keyphrase generation models
can be classified into
\textsc{One2One} \citep{meng-etal-2017-deep}, \textsc{One2Seq}  \citep{yuan-etal-2020-one} and \textsc{One2Set} \citep{DBLP:conf/acl/YeGL0Z20}, as shown in Figure \ref{fig:paradigm}.

\textbf{\textsc{One2One}.}
Typically, during model training, each training instance
contains an input document and only one corresponding keyphrase from the splitted target keyphrases.
During inference, 
\textsc{One2One} models adopt beam search to produce candidate phrases and then pick the top-\textit{K} ranked ones as the final keyphrases. 

As the earliest paradigm,
it has a far-reaching impact but neglects the correlation among keyphrases, limiting the potential of keyphrase generation models.

\textbf{\textsc{One2Seq}.}
To deal with the above issue, the \textsc{One2Seq} paradigm models keyphrase generation as a sequence generation task.
To this end,
target keyphrases are sorted in a predefined order
and concated as a sequence with delimiters.
Usually,
present keyphrases are firstly sorted according to their occurrence,
while absent keyphrases are then
randomly sorted \citep{DBLP:journals/corr/abs-1909-03590,DBLP:conf/naacl/MengYWZTH21}.

Due to the advantage of exploiting the semantic interdependence between keyphrases, 
\textsc{One2Seq} has become the most commonly-used paradigm.
However,
its premise of a predefined order introduces a bias into model training,
especially when the order of generated keyphrases is inconsistent with the predefined one.
Besides, \textsc{One2Seq} models tend to generate duplicated keyphrases \citep{DBLP:conf/acl/ChenCLK20, DBLP:conf/acl/YeGL0Z20}.


\textbf{\textsc{One2Set}.}
Furthermore,
to address the above bias defect of \textsc{One2Seq},
\cite{DBLP:conf/acl/YeGL0Z20} propose \textsc{One2Set}, 
where the keyphrase generation is modeled as a set generation task. 
Typically, 
its decoder utilizes 
different learnable control codes to generate a set of keyphrases in parallel.
During model training, the training loss is calculated according to the one-to-one alignments between the predicted keyphrases and target ones determined by the Hungarian Algorithm \citep{kuhn1955hungarian}.




\subsection{Document Encoding}
Typically,
CopyRNN \citep{meng-etal-2017-deep} adopts RNN as its encoder and thus suffers from low efficiency when handling long documents.
To solve this problem, \cite{DBLP:conf/icsai/ZhangFX17} replace RNN with CNN to boost encoding efficiency.

Besides, some researchers argue that sentences should be treated differently due to their unequal importance in document encoding.
\cite{DBLP:conf/aaai/ChenGZKL19} design Title-Guided Network, which additionally uses the title as a query to gather the information of title-relevant words in the input document.
\cite{DBLP:conf/sigir/KimSH21} takes into account useful structures of web documents such as title, body, header, query, to build a word graph representing both position-based proximity and structural relations.
\cite{DBLP:journals/corr/abs-2012-06754} use a selection network to filter unimportant sentences, while \cite{ahmad-etal-2021-select} apply this network to adjust the weights of the decoder copying mechanism.

Meanwhile,
researchers also focus on incorporating more information into the encoder.
For instance,
\cite{DBLP:conf/acl/ZhaoZ19} explore linguistic information for document encoding.
To alleviate data sparsity in social media, \cite{wang-etal-2019-topic-aware} apply a variational neural network to incorporate topic information into the model.

\subsection{Decoding Strategies}
Unlike the conventional decoder that 
can predict both present and absent keyphrases,
\cite{DBLP:conf/sigir/SunTDDN19} propose a diversified 
Pointer Network decoder for the \textsc{One2One} paradigm, which only copies a set of diverse present keyphrases.

Meanwhile,
more researchers focus on refining the decoding manners under \textsc{One2Seq} paradigm.
For example,
\cite{DBLP:conf/acl/ChenCLK20} 
propose an exclusive hierarchical decoder
that
involves two levels of decoding to exploit the phrase-level and word-level correlation
for keyphrase generation.
Similarly,
\cite{DBLP:conf/cikm/SantoshVVSD21}
model the above-mentioned hierarchical structure by
incorporating a conditional variational autoencoder.
Besides, \cite{DBLP:conf/sigir/ZhangJY0W22} propose
a hierarchical topic-guided variational neural network by
integrating the hierarchical topic information to guide the keyphrases generation.
Some researchers argue that
uniformly modeling the generation of present and absent keyphrases is unreasonable, since their prediction difficulties are significantly different.
\cite{DBLP:conf/naacl/ZhaoBWWHZ21} propose a Select-Guide-Generate decoding strategy, which firstly selects present keyphrases from the input document and then 
exploits these keyphrases
to guide the generation of absent ones.
Similarly,
\cite{DBLP:journals/taslp/LiuLW21} first fine-tune a BERT-based model to identify present keyphrases from the input document, and then utilize the BERT, which fully encodes the knowledge of present keyphrases, to benefit the generation of absent ones.
\cite{DBLP:conf/acl/WuLLNCZW21}
jointly train present keyphrase extraction and absent keyphrase generation, exploiting their mutual relation via stacker relation layer and bag-of-words constraints.
Very recently, \cite{wu2022fast} propose a mask-predict decoder to explore constrained and non-autoregressive generation for absent keyphrase generation.

\subsection{Model Training Strategies}
\cite{DBLP:conf/acl/ChanCWK19} propose a reinforcement learning (RL) approach with an adaptive reward for keyphrase generation.
If the model does not generate enough keyphrases, the reward is defined as the recall score that encourages the model to generate enough keyphrases.
Otherwise, the $F_1$ score is used as the reward to 
prevent the model from over-generating incorrect keyphrases.
To  ease the
synonym problem,
\cite{DBLP:conf/emnlp/Luo0YQZ21} further improve the RL reward function by considering
word-level $F_1$ score, edit distance,
duplication rate, and generation quantity.

Besides, researchers apply generative adversarial networks
to the keyphrase generation task  \citep{DBLP:conf/aaai/SwaminathanGZMG20, swaminathan-etal-2020-preliminary, lancioni-etal-2020-keyphrase},
where the generator is trained to produce accurate keyphrases and the discriminator is expected to distinguish machine-generated and human-curated keyphrases.

Many researchers apply multitask model to the keyphrase generation task \citep{DBLP:conf/naacl/ChenCLBK19, ahmad-etal-2021-select}.
Typically,
\cite{DBLP:conf/emnlp/YeW18} jointly train keyphrase generation and title generation to improve the generalization ability of the model.
Similarly, \cite{DBLP:conf/acl/ZhaoZ19} introduce POS tagging as an auxiliary task of keyphrase generation.


\subsection{Exploitation of External Information}
Inspired by the studies of other NLP tasks \citep{liu2018entity,wang2018ripplenet,zhang2019ernie}, researchers explore 
the information beyond input documents
to generate better keyphrases.

In this regard,
\cite{DBLP:journals/corr/abs-2004-09800} employ a cross-document attention to leverage similar documents for better document encoding.
\cite{DBLP:journals/corr/abs-2112-06776}  explore numerous ways to incorporate additional data for keyphrase generation and find that the summary of the article is the most beneficial.
Besides, researchers consider the keyphrases of similar documents.
\cite{DBLP:conf/naacl/ChenCLBK19} leverage the retrieved keyphrases from similar documents to guide the keyphrase generation and re-ranking.
\cite{DBLP:conf/pakdd/SantoshSBD21} also collect additional keyphrases from similar documents to automatically form a gazetteer,
which is used to 
enrich the vocabulary for
improving keyphrase generation.
To exploit both similar documents and their keyphrases,
\cite{DBLP:conf/emnlp/YeCGZ21} construct a heterogeneous keyword-document graph model, which is equipped with a reference-aware decoder to copy words from the input document and its similar ones.
To deal with the data without title,
\cite{DBLP:conf/emnlp/KimJCH21} construct a structure graph using the input document and its related but absent keyphrases retrieved from other documents.
This graph can
provide structure-aware representations for better keyphrase generation.
Besides,
\cite{wang-etal-2020-cross-media} utilize the rich features embedded in the matching images to explore the joint effects of texts and images for keyphrase prediction.


\subsection{Solving Duplication and Coverage Issues of Generated Keyphrases}
\cite{DBLP:conf/emnlp/ChenZ0YL18}
point out that the \textsc{One2One} paradigm neglects the correlation among keyphrases,
leading to duplication and coverage issues of generated keyphrases.
To solve these issues,
they propose CoryRNN that reviews preceding keyphrases to eliminate duplicates, and
utilizes the coverage mechanism \citep{tu-etal-2016-modeling} to improve the coverage for keyphrases.

The \textsc{One2Seq} paradigm has the same issues,
which become more serious when generating long keyphrase sequences.
To deal with this defect,
\cite{yuan-etal-2020-one} employ orthogonal regularization to
explicitly distinguish the delimiter-generated hidden states,
so as to improve the diversity of generated keyphrases.
\cite{DBLP:conf/coling/BahuleyanA20} use an unlikelihood training loss to produce diverse keyphrases.
Along this line,
\cite{DBLP:conf/acl/ChenCLK20} explore not only an training strategy with an exclusive loss,
but also an exclusive search strategy to avoid generating duplicate keyphrases.
In this way, the model is encouraged to generate keyphrases with different first words.


\subsection{Low-resource Keyphrase Generation}
The performance of keyphrase generation
models deeply depends on
the quantity and quality of training data.
Unfortunately,
the commonly-used labeled datasets
are often
relatively small,
making low-resource keyphrase generation a realistic and valuable research direction

\cite{DBLP:conf/emnlp/YeW18} propose a semi-supervised model 
that first generates pseudo keyphrases for unlabeled documents and then use them as incremental training data.
Besides,
\cite{DBLP:journals/corr/abs-2104-08729}
use unsupervised extraction models to collect keyphrases
and then draw
pesudo keyphrases for each document based on lexical and semantic level similarities.
Finally, the pesudo absent keyphrases
are used to train and update the model.

Recently, due to pre-trained models contain abundant knowledge that may benefit keyphrase generation,
keyphrase generation based on pre-trained models have received a rising interest.
In this respect,
\cite{DBLP:conf/acl/WuLLNCZW21} first introduce the pre-trained model UniLM \citep{DBLP:conf/nips/00040WWLWGZH19} into keyphrase generation.
Additionally,
\cite{DBLP:journals/corr/abs-2112-06776}
utilize Longformer \citep{DBLP:journals/corr/abs-2004-05150} to deal with the keyphrase generation for long documents.
Besides,
BART \citep{lewis-etal-2020-bart}, a denoising self-supervised autoencoder,
is extensively
applied due to its great potential in text generation tasks.
For instance,
\cite{DBLP:journals/corr/abs-2201-05302}
directly construct an \textsc{One2Seq} model based on the fine-tuned BART.
\cite{DBLP:journals/corr/abs-2112-08547} propose KeyBART,
which uses boundary tokens and position embeddings to predict the masked keyphrase
and then
determine whether a keyphrase is replaced or retained.
In addition to the above masked keyphrase prediction,
\cite{DBLP:journals/corr/abs-2203-08118} introduce salient span recovery to fine-tune BART for learning better intermediate representations.
\cite{wu2022fast} apply a prompt-based learning approach for constrained absent keyphrase generation.
They firstly define overlapping words between absent keyphrase and document as keywords,
and then use a mask-predict decoder to generate the final absent keyphrase under the constraints of prompt.


\subsection{Keyphrase Ranking}
Due to the property of beam search,
\textsc{One2One} models tend
to select short phrases.
To deal with this issue,
\cite{DBLP:journals/corr/abs-1909-09485} introduce word-level and ngram-level attention scores to boost the ranking scores of long keyphrases.
Besides, \cite{DBLP:journals/corr/abs-2104-08729} 
combine the TF-IDF relatedness and embedding-based keyphrase-document cosine similarity to rank phrases.
When reranking phrases, \cite{DBLP:conf/naacl/ChenCLBK19} also consider phrases retrieved from similar documents
and phrases extracted from documents.
In addition,
\cite{DBLP:conf/emnlp/YeW18}
apply beam search into an \textsc{One2Seq} paradigm based model, which generates multiple candidate phrase sequences and then
collect unique keyphrases from the top-ranked beams in descending order.


\section{Datasets}

\begin{table}[htbp]
\footnotesize
\centering
\vskip 2mm
\setlength{\tabcolsep}{3mm}{
\caption{The commonly-used datasets for keyphrase predictions.}
    \centering
    \begin{tabular}{l|ccc}
    \toprule
        {\bf Dataset} & {\bf Domain} & {\bf Language} & {\bf Docs} \\ \hline
        Inspec \citep{hulth-2003-improved} & Papers & EN & 2.0K  \\ 
        NUS \citep{DBLP:conf/icadl/NguyenK07} & Papers & EN & 211  \\
        PubMed \citep{schutz2008keyphrase} & Papers & EN & 1.3K \\
        Krapivin \citep{krapivin2009large} & Papers & EN & 2.3K  \\ 
        Citeulike-180 \citep{DBLP:conf/emnlp/MedelyanFW09} & Papers & EN & 181 \\
        SemEval-2010 \citep{DBLP:conf/semeval/KimMKB10} & Papers & EN & 244  \\ 
        TALN \citep{DBLP:conf/taln/Boudin13} & Papers & EN/FR & 521/1.2K \\
        KDD \citep{DBLP:conf/aaai/GollapalliC14} & Papers & EN & 755 \\
        WWW \citep{DBLP:conf/aaai/GollapalliC14} & Papers & EN & 1.3K \\
        TermLTH-Eval \citep{bougouin-etal-2016-termith} & Papers & FR & 400 \\
        KP20k \citep{meng-etal-2017-deep} & Papers & EN & 567.8K  \\
        LDPK3K \citep{DBLP:journals/corr/abs-2203-15349} & Papers & EN & 96.8K \\
        LDPK10K \citep{DBLP:journals/corr/abs-2203-15349} & Papers & EN & 1.3M \\
        
        DUC \citep{DBLP:conf/aaai/WanX08} & News & EN & 308  \\ 
        110-PT-BN-KP \citep{DBLP:conf/interspeech/MarujoVN11} & News & PT & 110 \\
        500N-KPCrowd \citep{marujo-etal-2012-supervised} & News & EN & 500 \\
        Wikinews \citep{DBLP:conf/ijcnlp/BougouinBD13} & News & FR & 100 \\
        PerKey \citep{DBLP:conf/istel/DoostmohammadiB18} & News & PER & 553.1K  \\ 
        KPTimes \citep{DBLP:conf/inlg/GallinaBD19} & News & EN & 279.9K  \\
        
        Twitter \citep{DBLP:conf/emnlp/ZhangWGH16} & Tweets & EN & 112.5K  \\
        Weibo \citep{wang-etal-2019-topic-aware} & Tweets & ZH & 46.3K \\
        Text-Image Tweets \citep{wang-etal-2020-cross-media} & Tweets & EN &  53.7K \\
        
        NZDL \citep{DBLP:conf/dl/WittenPFGN99} & Reports & EN & 1.8K  \\
        Blogs \citep{grineva2009extracting} & Web pages & EN & 252 \\
        StackExchange \citep{wang-etal-2019-topic-aware} & QA & EN & 49.4K \\
    \bottomrule
    \end{tabular}
    \vskip -1mm
\label{used_dataset}}
\end{table}

The commonly-used datasets for keyphrase prediction are shown in Table~\ref{used_dataset}.
According to domains,
they could be divided into reports, News, tweets, web pages, QA and scientific articles.
Most of these datasets are in English,
a few are in French, Persian, Chinese, and Portuguese.



As the most widely-used dataset,
KP20k consists of
articles in computer science from various online digital libraries.
Overall, these datasets are relatively small, which is not applicable to industrial applications.
Hence, it is urgent to construct large-quantity and high-quality multilingual datasets, so as to further promote the development of keyphrases prediction.

Considering the tradeoff between cost and quality of expert annotations,
\cite{chau-etal-2020-understanding} explore multiple annotation strategies, including self review, peer review, and so on.

\section{Evaluation Metrics}

Let $\hat{Y}$=$(\hat{y}_1, \hat{y}_2, ..., \hat{y}_m)$
and
$Y$=$(y_2, y_2, ..., y_n)$
to be the predicted and target keyphrases, respectively.
The common practice is to use only top $k$ predictions with the highest scores for evaluation,
where $k$ is a pre-defined constant (usually 5 or 10).
Particularly, to eliminate the influence of morphology,
the predicted keyphrases are stemmed
by applying Porter Stemmer\footnote{\url{https://github.com/nltk/nltk/blob/develop/nltk/stem/porter.py}} \citep{meng-etal-2017-deep}.

The commonly-used metrics include precision, recall and $F_1$ scores.
Early studies use $F_1@5$ and $F_1@10$ to evaluate the quality of generated present keyphrases, and $R@5$ and $R@10$ to measure the quality of generated absent keyphrases \citep{meng-etal-2017-deep, DBLP:conf/emnlp/ChenZ0YL18, DBLP:conf/aaai/ChenGZKL19}.
Formally, these metrics are defined as follows:

\begin{equation}
P@k = \frac{|\hat{Y}_{:k}\cap Y|}{|\hat{Y}_{:k}|},
\end{equation}
representing the correct proportion of keyphrases in predictions.
\begin{equation}
R@k = \frac{|\hat{Y}_{:k}\cap Y|}{|Y|},
\end{equation}
measuring the correct rate of the predicted keyphrase in references.
\begin{equation}
F_1@k = \frac{2\* P@k \* R@k}{P@k+R@k}, 
\end{equation}
which is a tradeoff between $P@k$ and $R@k$.





Considering the fact that a model often predicts varying numbers of keyphrases,
\cite{yuan-etal-2020-one} argue that
the metrics with the pre-defined constant $k$
cannot accurately evaluate the quality of predicted keyphrases.
Thus,
they extend $F_1@k$ to two metrics: 1) $F_1@O$:
this metric sets $k$ as the number of target keyphrases instead of a pre-defined constant;
2) $F_1@M$: this metric takes all predictions into account.
Futhermore,
\cite{DBLP:conf/acl/ChanCWK19} improve {\emph{$F_1@M$}} by filling target keyphrases with blanks when their number is less than the number of predicted keyphrases.

However, conventional metrics, such as \emph{$F_1$}, 
which assess the prediction quality at the phrase level, 
do not take into account the partially matched predictions.
To deal with this issue,
\cite{DBLP:conf/emnlp/Luo0YQZ21} propose Fine-Grained ($FG$) evaluation score that considers prediction orders and qualities at the token level, and prediction diversity and numbers at the instance level.




Besides,
\cite{habibi2013diverse}
introduce \emph{$\alpha$-nDCG} \citep{clarke2008novelty}
to measure the diversity of predicted keyphrases, 
where \emph{nDCG} represents Normalized Discounted
Cumulative Gain measure \citep{DBLP:journals/tois/JarvelinK02} and the parameter $\alpha$ is a trade-off between relevance and diversity.
\cite{DBLP:conf/acl/ChanCWK19} measure
the mean absolute error (\emph{MAE}) between the number of predicted keyphrases and the number of target keyphrases. 
In addition,
\cite{DBLP:conf/acl/ChenCLK20} define \emph{DupRatio} 
to evaluate the duplication rate of the predicted keyphrases.

\subsection{Implementation Details}

In the experiments of keyphrase extraction,
we consider the following typical unsupervised models:
statistical models including TF-IDF \citep{DBLP:journals/ipm/SaltonB88}, YAKE \citep{DBLP:conf/ecir/0001MPJNJ18a}, graph-based models consisting of TextRank \citep{DBLP:conf/emnlp/MihalceaT04}, SingleRank \citep{DBLP:conf/aaai/WanX08}, TopicRank \citep{DBLP:conf/ijcnlp/BougouinBD13}, PositionRank \citep{florescu-caragea-2017-positionrank}, MultipartieRank \citep{boudin-2018-unsupervised}, and deep learning-based models such as EmbedRank \citep{DBLP:journals/corr/abs-1801-04470}, SIFRank \citep{DBLP:journals/access/SunQZWZ20}, SIFRank+ \citep{DBLP:journals/access/SunQZWZ20}, UKERank \citep{DBLP:conf/emnlp/LiangW0L21},
and JointKPE \citep{DBLP:conf/nlpcc/SunLXLB21}. Please Note that JointKPE is the current SOTA supervised model.
In the experiments of keyphrase generation,
we compare the typical generation models under three kinds of common paradigms:
1) \textsc{One2One}, CopyRNN \citep{meng-etal-2017-deep} and KG-KE-KR-M \citep{DBLP:conf/naacl/ChenCLBK19},
2) \textsc{One2Seq}, CatSeq \citep{yuan-etal-2020-one}, catSeqTG-2RF$_1$ \citep{DBLP:conf/acl/ChanCWK19} and Transformer \citep{DBLP:conf/acl/YeGL0Z20}, and
3) \textsc{One2Set}, SetTrans \citep{DBLP:conf/acl/YeGL0Z20} and WR-SetTrans\citep{DBLP:conf/emnlp/XieWYLXWZS22}.
Besides, we compare the performance of large language models in keyphrase prediction, including BART \citep{lewis-etal-2020-bart}, T5 \citep{DBLP:journals/jmlr/RaffelSRLNMZLL20}, KeyBART \citep{DBLP:journals/corr/abs-2112-08547} and ChatGPT\footnote{\url{https://chat.openai.com/chat}}.

During model training, we strictly use the same experiment settings as their original papers.
For the model involving multiple variants,
we only report the performance of its variant with the best performance.
Particularly, following \cite{yuan-etal-2020-one}, we use two experimental settings for \textsc{One2Seq} paradigm models.
When using ChatGPT, we explore three commonly-used settings, including zero-shot\footnote{We use the official released prompt (\url{https://platform.openai.com/examples/default-keywords}) for keyphrase prediction.}, 1-shot, and 5-shot. Specifically, we retrieve the most relevant training instances for the given input document according to the cosine distance of the MiniLM \citep{DBLP:conf/nips/WangW0B0020} embedding. These pertinent training instances are concatenated at the beginning of the input document and then fed into the ChatGPT to obtain the ultimate predictions for keyphrases.
Particularly, to alleviate the instability of neural networks, 
we run the generation models for 3 times with different seeds and report the average results.
Finally, we evaluate the present keyphrase and absent keyphrase predictions, respectively.

\begin{table*}[t]
\centering
\setlength{\tabcolsep}{0.2mm}{
\caption{Results of present keyphrase prediction using extraction models. To ensure fair comparsions, we only use the target present keyphrase to evaluate the performance of extraction models, while the previous studies use all target keyphrases.}\label{table:present-extraction-main-performance}}
\begin{threeparttable}[width=0.5\textwidth]
    \begin{tabular}{p{6.0cm}|rr|rr|rr|rr|rr}
    \toprule
    \multicolumn{1}{c|}{\multirow{2}[2]{*}{\textbf{Model}}} & \multicolumn{2}{c|}{\textbf{Inspec}} & \multicolumn{2}{c|}{\textbf{NUS}} & \multicolumn{2}{c|}{\textbf{Krapivin}} & \multicolumn{2}{c|}{\textbf{SemEval}} & \multicolumn{2}{c}{\textbf{KP20k}} \\
          & \multicolumn{1}{l}{F1@5} & \multicolumn{1}{l|}{F1@M} & \multicolumn{1}{l}{F1@5} & \multicolumn{1}{l|}{F1@M} & \multicolumn{1}{l}{F1@5} & \multicolumn{1}{l|}{F1@M} & \multicolumn{1}{l}{F1@5} & \multicolumn{1}{l|}{F1@M} & \multicolumn{1}{l}{F1@5} & \multicolumn{1}{l}{F1@M} \\
    \midrule
    \multicolumn{11}{c}{Unsupervised Statistical Extraction Models} \\
    \midrule
    TF-IDF \citep{DBLP:journals/ipm/SaltonB88} & 0.132  & 0.175  & 0.214  & 0.213  & 0.145  & 0.131  & 0.151  & 0.190  & 0.172  & 0.146  \\
    YAKE \citep{DBLP:conf/ecir/0001MPJNJ18a}  & 0.183  & 0.193  & 0.221  & 0.212  & 0.188  & 0.131  & 0.202  & 0.204  & 0.189  & 0.145  \\
     \midrule
    \multicolumn{11}{c}{Unsupervised Graph-based Extraction Models} \\
    \midrule
    TextRank \citep{DBLP:conf/emnlp/MihalceaT04} & 0.321  & 0.363  & 0.092  & 0.169  & 0.118  & 0.144  & 0.093  & 0.200  & 0.091  & 0.120  \\
    SingleRank \citep{DBLP:conf/aaai/WanX08} & 0.325  & 0.362  & 0.151  & 0.195  & 0.152  & 0.147  & 0.146  & 0.212  & 0.134  & 0.131  \\
    TopicRank \citep{DBLP:conf/ijcnlp/BougouinBD13} & 0.266  & 0.301  & 0.210  & 0.154  & 0.168  & 0.118  & 0.201  & 0.163  & 0.167  & 0.114  \\
    PositionRank \citep{florescu-caragea-2017-positionrank} & 0.306  & 0.338  & 0.228  & 0.208  & 0.186  & 0.143  & 0.245  & 0.229  & 0.183  & 0.138  \\
    MultipartiteRank \citep{boudin-2018-unsupervised} & 0.269  & 0.322  & 0.244  & 0.188  & 0.181  & 0.132  & 0.227  & 0.206  & 0.185  & 0.132  \\
    \midrule
    \multicolumn{11}{c}{Unsupervised Deep Learning-based Extraction Models} \\
    \midrule
    EmbedRank \citep{DBLP:journals/corr/abs-1801-04470} & 0.333  & 0.376  & 0.166  & 0.199  & 0.167  & 0.150  & 0.185  & 0.233  & 0.153  & 0.135  \\
    SIFRank \citep{DBLP:journals/access/SunQZWZ20} & \textbf{0.368}  & \textbf{0.385} & 0.143  & 0.193  & 0.164  & 0.151  & 0.165  & 0.213  & 0.138  & 0.133  \\
    SIFRank+ \citep{DBLP:journals/access/SunQZWZ20} & 0.348  & 0.384  & 0.246  & 0.203  & 0.194  & 0.153  & 0.244  & 0.223  & 0.195  & 0.138  \\
    UKERank \citep{DBLP:conf/emnlp/LiangW0L21} & 0.350  & 0.384  & 0.238  & 0.202  & 0.187  & 0.162  & 0.250  & 0.228  & 0.178  & 0.138  \\
    \midrule
    \multicolumn{11}{c}{Supervised Deep Learning-based Extraction Models} \\
    \midrule
    Sequence Tagging(Roberta-base) \citep{DBLP:conf/nlpcc/SunLXLB21} &  0.331     &   0.336    &   0.321    &    0.177   &    \textbf{0.476}   &    \textbf{0.319}   &    0.379   &   0.291    &   0.416    &  \textbf{0.240} \\
    JointKPE \citep{DBLP:conf/nlpcc/SunLXLB21} &  0.352     &   0.348    &   \textbf{0.476}    &   \textbf{0.335}    &   0.360    &   0.202    &  \textbf{0.393}     &   \textbf{0.306}    &   \textbf{0.417}    &  0.239 \\
\bottomrule
\end{tabular}
\end{threeparttable}
\end{table*}

\begin{table*}[ht]

\centering
\setlength{\tabcolsep}{0.3mm}{
\caption{Results of present keyphrase prediction using generation models. \#\textit{bs}~denotes beam size. $^\dag$~indicates previously reported scores.}\label{table:present-generative-performance}}
\begin{threeparttable}[width=0.8\textwidth]
    \begin{tabular}{l|rr|rr|rr|rr|rr}
    \toprule
    \multicolumn{1}{c|}{\multirow{2}[2]{*}{\textbf{Model}}} & \multicolumn{2}{c|}{\textbf{Inspec}} & \multicolumn{2}{c|}{\textbf{NUS}} & \multicolumn{2}{c|}{\textbf{Krapivin}} & \multicolumn{2}{c|}{\textbf{SemEval}} & \multicolumn{2}{c}{\textbf{KP20k}} \\
          & \multicolumn{1}{l}{F1@5} & \multicolumn{1}{l|}{F1@M} & \multicolumn{1}{l}{F1@5} & \multicolumn{1}{l|}{F1@M} & \multicolumn{1}{l}{F1@5} & \multicolumn{1}{l|}{F1@M} & \multicolumn{1}{l}{F1@5} & \multicolumn{1}{l|}{F1@M} & \multicolumn{1}{l}{F1@5} & \multicolumn{1}{l}{F1@M} \\
    \midrule
    \multicolumn{11}{c}{\textsc{One2One} Paradigm-based Models} \\
    \midrule
    CopyRNN(\#\textit{bs}=200) \citep{meng-etal-2017-deep} & 0.272 & 0.293 & 0.356 & 0.306 & 0.283 & 0.214 & 0.294 & 0.257 & 0.336 & 0.255 \\
    KG-KE-KR-M(\#\textit{bs}=200) \citep{DBLP:conf/naacl/ChenCLBK19}
    & 0.324   &   0.362    &    0.421  &   0.342    &  0.304     &    0.273   &   0.325    &    0.293   &   \textbf{0.400}    & 0.277 \\
    \midrule
    \multicolumn{11}{c}{\textsc{One2Seq} Paradigm-based Models} \\
    \midrule
    CatSeq(\#\textit{bs}=1) \citep{yuan-etal-2020-one} & 0.229  & 0.266  & 0.324  & 0.394  & 0.270  & 0.344  & 0.245  & 0.296  & 0.292  & 0.365  \\
    CatSeq(\#\textit{bs}=50) \citep{yuan-etal-2020-one} & 0.328  & \textbf{0.398} & 0.417 & 0.395 & 0.352 & 0.316 & \textbf{0.343} & 0.334 & 0.360 & 0.302 \\
    catSeqTG-2RF$_1$(\#\textit{bs}=1) \citep{DBLP:conf/acl/ChanCWK19}
    & 0.253 & 0.301 & 0.375 & 0.433 & 0.300 & 0.369 & 0.287 & 0.329 & 0.321 & 0.386  \\
    
    
    Transformer(\#\textit{bs}=1) \citep{DBLP:conf/acl/YeGL0Z20} & 0.285  & 0.331  & 0.371  & 0.418  & 0.308  & 0.356  & 0.287  & 0.319  & 0.330  & 0.373  \\
    
    \midrule
    \multicolumn{11}{c}{\textsc{One2Set} Paradigm-based Models} \\
    \midrule
    SetTrans(\#\textit{bs}=1) \citep{DBLP:conf/acl/YeGL0Z20} & 0.281  & 0.318  & 0.406  & \textbf{0.452}  & 0.339  & \textbf{0.374 } & 0.322  & 0.354 & 0.354  & \textbf{0.390 } \\
    WR-SetTrans(\#\textit{bs}=1) \citep{DBLP:conf/emnlp/XieWYLXWZS22} & \textbf{0.330}  & 0.351  & \textbf{0.428} & \textbf{0.452}  & \textbf{0.360}  & 0.362 & \textbf{0.360}  & \textbf{0.370} & 0.370 & 0.378 \\
\bottomrule
\end{tabular}
\end{threeparttable}
\end{table*}
\begin{table*}[ht]

\centering
\setlength{\tabcolsep}{0.5mm}{
\caption{Results of absent keyphrase prediction using generation models.}

\begin{threeparttable}[width=0.8\textwidth]
    \begin{tabular}{l|rr|rr|rr|rr|rr}
    \toprule
    \multicolumn{1}{c|}{\multirow{2}[1]{*}{\textbf{Model}}} & \multicolumn{2}{c|}{\textbf{Inspec}} & \multicolumn{2}{c|}{\textbf{NUS}} & \multicolumn{2}{c|}{\textbf{Krapivin}} & \multicolumn{2}{c|}{\textbf{SemEval}} & \multicolumn{2}{c}{\textbf{KP20k}} \\
          & \multicolumn{1}{l}{F1@5} & \multicolumn{1}{l|}{F1@M} & \multicolumn{1}{l}{F1@5} & 
          \multicolumn{1}{l|}{F1@M} & \multicolumn{1}{l}{F1@5} & 
          \multicolumn{1}{l|}{F1@M} & \multicolumn{1}{l}{F1@5} & 
          \multicolumn{1}{l|}{F1@M} & \multicolumn{1}{l}{F1@5} & 
          \multicolumn{1}{l}{F1@M} \\
    \midrule
    \multicolumn{11}{c}{\textsc{One2One} Paradigm-based Models} \\
    \midrule
    CopyRNN(\#\textit{bs}=200) \citep{meng-etal-2017-deep}  &  0.007 &  0.007 & 0.009  & 0.012  & 0.013  & 0.019  &  0.007 & 0.011  & 0.011  & 0.013  \\
    KG-KE-KR-M(\#\textit{bs}=200) \citep{DBLP:conf/naacl/ChenCLBK19}$^\dag$ &  0.024  &   0.028    &    \textbf{0.060}  &   \textbf{0.076}    &   \textbf{0.059}   &    0.063   &   0.031  &   0.040  &   \textbf{0.070}    & \textbf{0.083} \\
    \midrule
    \multicolumn{11}{c}{\textsc{One2Seq} Paradigm-based Models} \\
    \midrule
    CatSeq(\#\textit{bs}=1) \citep{yuan-etal-2020-one} & 0.005  & 0.009  & 0.015  & 0.026  & 0.018  & 0.034  & 0.015  & 0.022  & 0.014  & 0.030  \\
    CatSeq(\#\textit{b}=50) \citep{yuan-etal-2020-one} &  0.021 & 0.028 & 0.038 & 0.052 & 0.051 & 0.065 & 0.030 & 0.038 & 0.041 & 0.058  \\
    catSeqTG-2RF$_1$(\#\textit{bs}=1) \citep{DBLP:conf/acl/ChanCWK19} & 0.012 & 0.021 & 0.019 & 0.031 & 0.030 & 0.053 & 0.021 & 0.030 & 0.027 & 0.050  \\
    Transformer(\#\textit{bs}=1) \citep{DBLP:conf/acl/YeGL0Z20} & 0.008  & 0.017  & 0.028  & 0.050  & 0.030  & 0.055  & 0.016  & 0.022  & 0.021  & 0.043  \\
    \midrule
    \multicolumn{11}{c}{\textsc{One2Set} Paradigm-based Models} \\
    \midrule
    SetTrans(\#\textit{bs}=1) \citep{DBLP:conf/acl/YeGL0Z20}  & 0.018  & 0.029 & 0.041 & 0.061  & 0.046 & 0.073 & 0.029 & 0.035 & 0.035 & 0.056 \\
     WR-SetTrans(\#\textit{bs}=1) \citep{DBLP:conf/emnlp/XieWYLXWZS22} & \textbf{0.025}  & \textbf{0.034}  & 0.057  & 0.071  & 0.057  & \textbf{0.074} &\textbf{ 0.040}  & \textbf{0.043 }& 0.050 & 0.064 \\
\bottomrule
\end{tabular}
\end{threeparttable}
\label{table:absent-main-performance}
}
\end{table*}

\section{Comparison between Existing Models}

To better understand advantages and disadvantages of different models,
we conduct several groups of experiments to compare representative models in different settings.
To this end, we use the KP20k training set to train various models, and then apply the same script\footnote{\url{https://github.com/kenchan0226/keyphrase-generation-rl/blob/master/evaluate_prediction.py}} to evaluate the model predictions on five commonly-used test sets: Inspec, NUS, Krapivin, SemEval, and KP20k.

\subsection{Comparison of Extraction Models}
The performance of extraction models is reported in Table~\ref{table:present-extraction-main-performance}.
Note that the previous studies in this aspect report the evaluation scores with respect to all target keyphrases.
To ensure fair comparisons, we only use the present keyphrases to evaluate the performance of various extraction models.

Overall, unsupervised statistical extraction models perform worst in this setting, 
and unsupervised graph-based extraction models surpass statistical ones.
This result is not surprising, because unsupervised graph-based extraction models not only use statistical features but also employ effective graph algorithms, such as clustering, graph propagation, etc. 
Moreover, due to the advantage of semantic representation learning,
deep learning-based models achieve the best result, 
echoing the development trend of natural language processing studies from statistical models to deep learning-based models.

\begin{figure*}[t]
\centering
    \begin{minipage}{0.8\textwidth}
        \centering
        \includegraphics[width=0.7\linewidth]{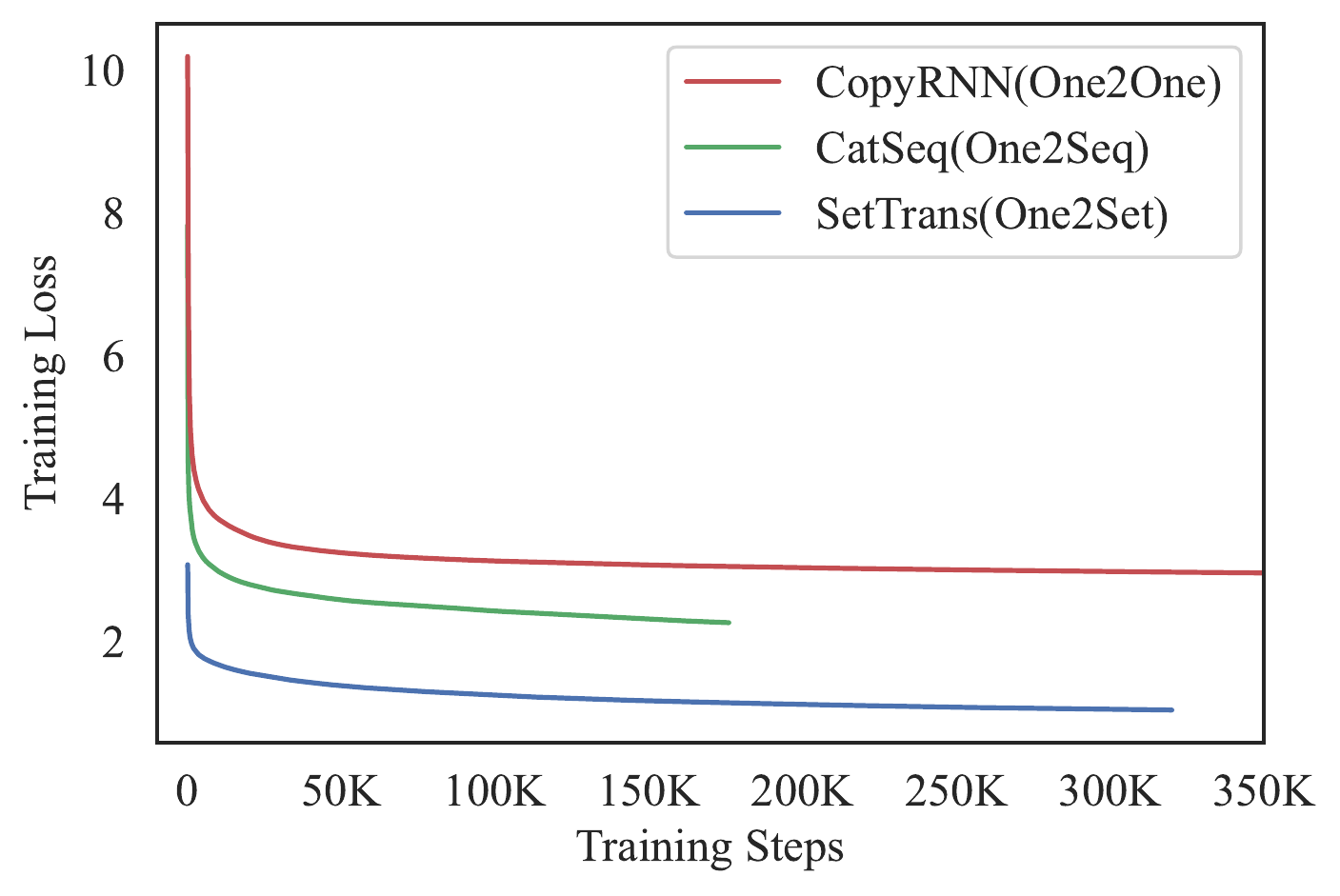}
        \caption{The training losses of representative models under three paradigms.}
        \label{fig:training loss}
    \end{minipage}
\end{figure*}

Besides, comparing unsupervised and supervised extraction models,
we can observe that supervised extraction models outperform unsupervised ones on most test sets except Inspec.  
Further analysis on Inspec will be provided in Section~\ref{generative-methods}.

\subsection{Comparison of Gneration Models}
\label{generative-methods}

\paragraph{Three Training Paradigms}

Figure~\ref{fig:training loss} shows the training losses of CopyRNN, CatSeq and SetTrans, which are the representative models under three paradigms.
CopyRNN suffers from the highest loss,
due to the difficulty of model training brought by the One2One paradigm where one input corresponds to multiple targets.
One2Seq paradigm alleviates the problem of inconsistent training instances by concatenating target keyphrases into a sequence and Cat2Seq achieves a relatively lower loss than CopyRNN.
Among three representative models, SetTrans has the lowest loss after convergence, demonstrating the advantage of the ONE2SET paradigm.
\begin{figure*}[hbpt]
	\centering

        \centering
        \includegraphics[width=0.7\textwidth]{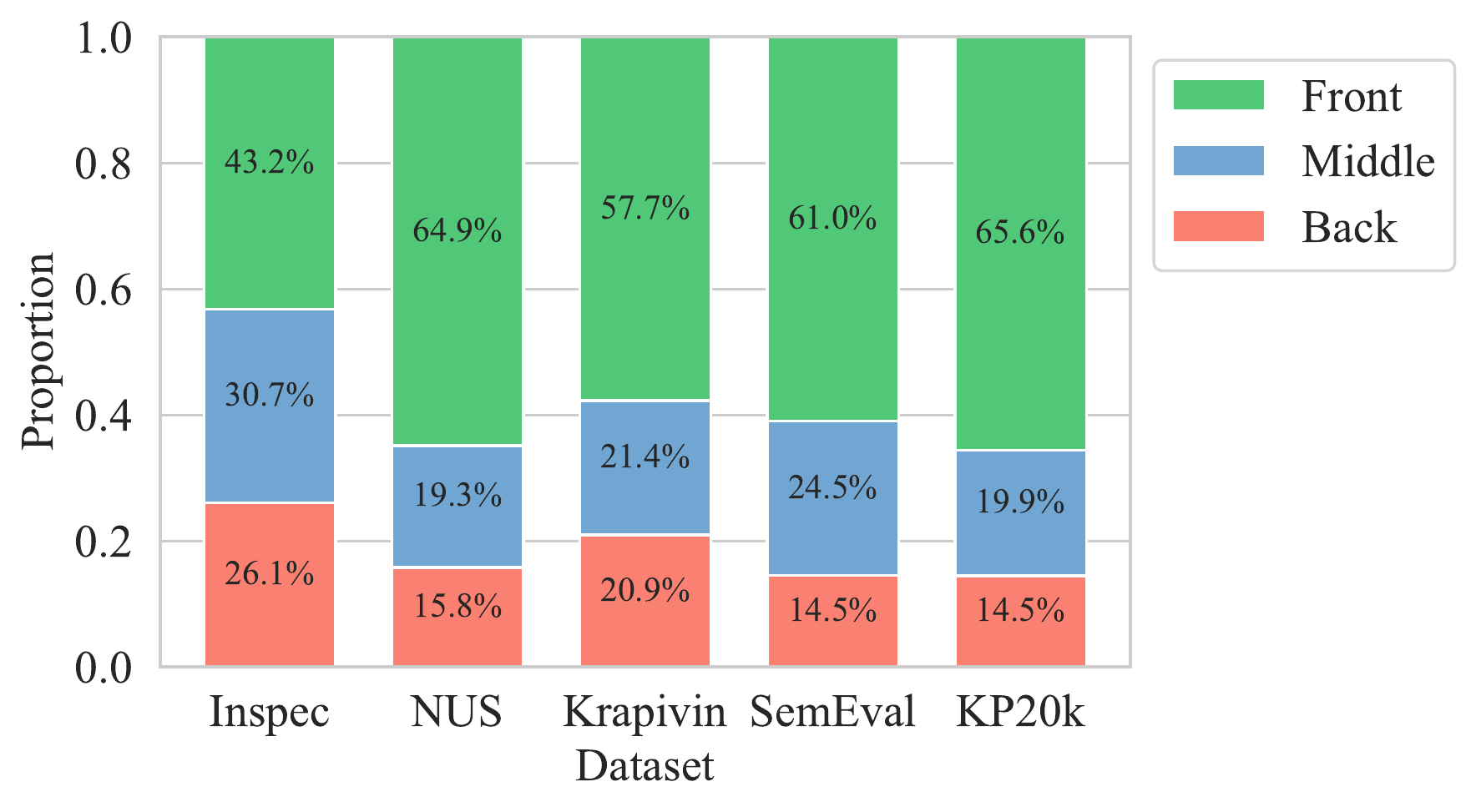}
	
	\centering
	\caption{The first occurrence position distribution of present keyphrases in input documents}
    \label{dataset_pos}
\end{figure*}

\paragraph{Comparison of SOTA Extraction Model and Generation Models}
From the last rows of Table~\ref{table:present-extraction-main-performance} and Table~\ref{table:present-generative-performance},
we observe that JointKPE \citep{DBLP:conf/nlpcc/SunLXLB21} outperfoms all generation models in terms of \emph{$F_1@5$}.
However, extraction models cannot dynamically decide the number of extracted keyphrases.
If the pre-defined number of extracted keyphrases is larger than the actual number of target keyphrases, it may introduce noise into the extracted phrases, resulting in a low \emph{$F_1@M$}.
Worse still, extraction models are unable to deal with the predictions of absent keyphrases, which account for a large proportion of target keyphrases.
Therefore, we argue that a combination of extraction and generation model, such as KG-KE-KR-M \citep{DBLP:conf/naacl/ChenCLBK19}, has the potential to achieve better overall results than single-mode models.

\begin{table*}[htbp]
  \centering
  \caption{Statistical features of five datasets.}
    \begin{tabular}{c|cccccc}
    \toprule
    Dataset & \#pre. KP/doc & \#abs. KP/doc & \#token/doc & Length of pre. KP & Length of abs. KP & \#doc \\
    \midrule
    Inspec & 7.23  & 2.59  & 134.10  & 2.44  & 2.72  & 500 \\
    NUS   & 6.34  & 5.31  & 230.13  & 1.95  & 2.56  & 211 \\
    Krapivin & 3.26  & 2.59  & 189.32  & 2.16  & 2.29  & 400 \\
    SemEval & 6.25  & 8.41  & 245.89  & 2.08  & 2.61  & 100 \\
    KP20k & 3.24  & 2.84  & 179.02  & 1.85  & 2.55 & 570, 802 \\
    \bottomrule
    \end{tabular}%
  \label{tab:statistic-datasets}%
\end{table*}%

\begin{figure}[hbpt]
	\centering
	\vspace{-0.15in}
	
	
			\includegraphics[width=0.6\linewidth]{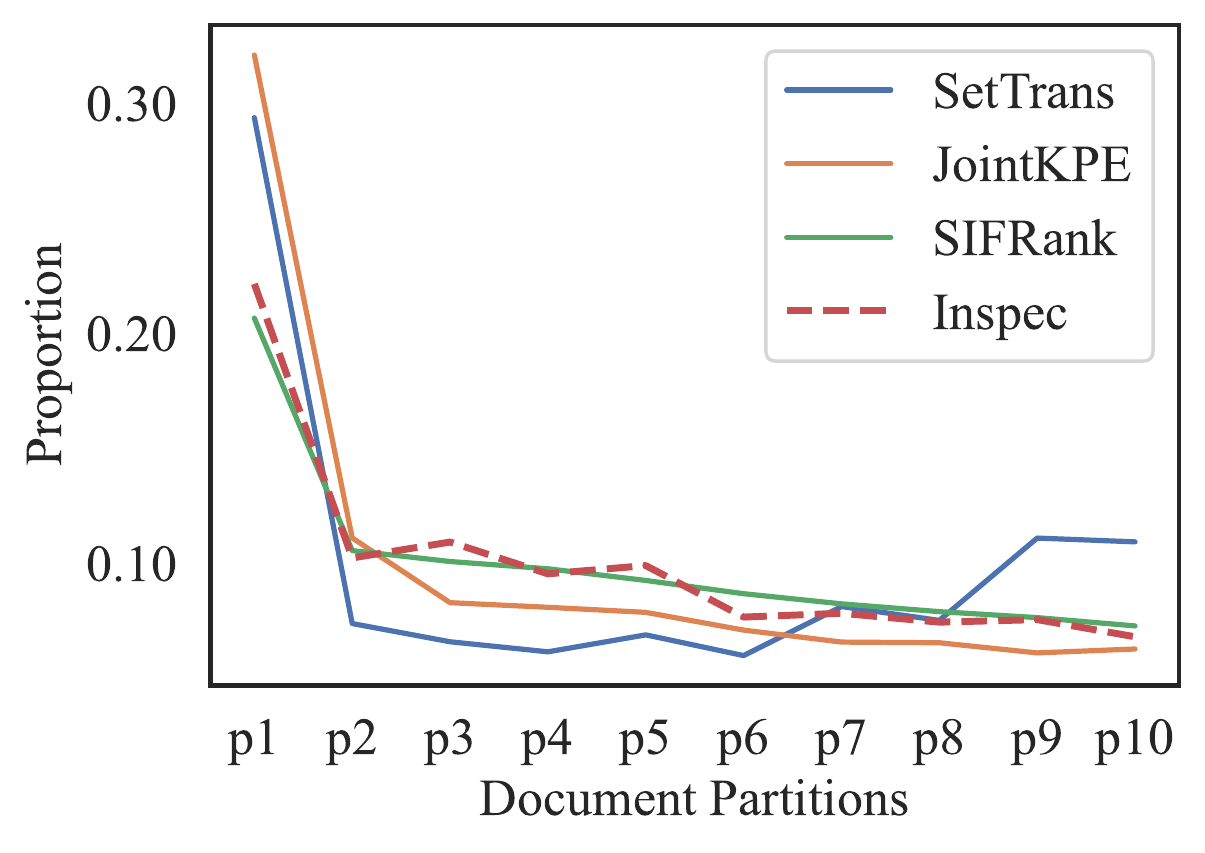}
	\caption{The first occurrence position distributions of the target present keyphrases and present keyphrases predicted by SIFRank, JointKPE and SetTrans in Inspec. The document has been divided into ten equal parts, and the x-axis indicates the index of the divided sub-document, for example, x = p1 means that the first 10\% of the document. The y-axis is the proportion of the keyphrase in this sub-document.}
	\vspace{-0.2in}		
	\label{settrans_sifrank}
\end{figure}

Back to Table~\ref{table:present-generative-performance},
KG-KE-KR-M performs significantly better than CopyRNN, proving the superiority of combing generation and extraction. 
Note that although KG-KE-KR-M incorporates retrieval and reranking techniques into \textsc{One2One} paradigm,
SetTrans\citep{DBLP:conf/acl/YeGL0Z20} still outperforms KG-KE-KR-M and other generation models in F1@M without special techniques, showing its advantages in predicting the keyphrase number for documents.

\paragraph{Analysis of the Inspec Dataset}

From Table~\ref{table:present-extraction-main-performance} and Table~\ref{table:present-generative-performance}, we observe that unsupervised deep learning-based extraction models achieve comparable or better performance than supervised deep learning-based extraction and generation models when predicting present keyphrases on Inspec.
To explain this phenomena, we further conduct the following analyses:

1) Table~\ref{tab:statistic-datasets} shows the statistical features of datasets.
Compared with other test sets, Inspec has the shortest average document length, the longest average length of present keyphrase, and the maximum number of present keyphrase, indicating Inspec is more suitable for extraction models than other datasets.

2) In Figure \ref{dataset_pos}, we also visualize the occurrence position distributions of present keyphrases in each dataset.
It reveals that the present keyphrases of the KP20k training set tend to occur in the front of the document.
This phenomenon becomes even more evident when analyzing the first occurrence positions of present keyphases.
As a result, the supervised models trained on KP20k tend to predict present keyphrases from the front of the document, which, however, is not applicable for Inspec, of which present keyphrases distribute evenly in the document.

3) Figure~\ref{settrans_sifrank} depicts the position distributions of the target present keyphrases and present keyphrases predicted by SIFRank, JointKPE, and SetTrans in Inspec.
Please note that they are the best unsupervised keyphrase extraction,
supervised keyphrase extraction and keyphrase generation models, respectively.
The distribution of present keyphrases predicted by SIFRank is very close to the distribution of Inspec, while other supervised models are quite different. 
It supports our hypothesis that supervised models, are deeply affected by the occurrence position distribution of keyphrases in training data, which leads to the degradation of model performance when the test set is domain-mismatch with the training data.

\subsection{Comparison of LLMs}
Recently, large language models (LLMs) have achieved remarkable success in various NLP tasks and have displayed a variety of capabilities. To evaluate the keyphrase prediction ability of these models, we compare the performance of commonly-used LLMs, including BART, T5, KeyBART, and ChatGPT, across five benchmark datasets.

Table~\ref{table:llm-present-generative-performance} and Table~\ref{table:llm-absent-main-performance} reports the experimental results. We find that compared with previous SOTA models, such as CatSeq, Transformer, SetTrans and WR-SetTrans, LLMs show modest improvements in both present and absent keyphrase predictions. Additionally, our comparison of SetTrans and LLMs suggests that the impact of increasing model parameters is overshadowed by the adoption of new training and inference paradigms.

By synthesizing all results of Table~\ref{table:present-extraction-main-performance}, Table~\ref{table:llm-present-generative-performance} and Table~\ref{table:llm-absent-main-performance},
we conclude that ChatGPT outperforms all other unsupervised keyphrase extraction methods in terms of F1@5-score and F1@M-score under the zero-shot setting, but is still inferior to the existing SOTA supervised models on almost all datasets. With more training instances,
the prediction ability of ChatGPT for both present and absent keyphrases can be significantly improved. Its superior performance on multiple benchmark datasets highlights its significance for practical applications in various domains. Further research could explore the more effective use of ChatGPT to fully exert its potential.

\begin{table*}[ht]
\centering
\setlength{\tabcolsep}{0.3mm}{
\caption{Results of present keyphrase prediction using large language models. \#\textit{bs}~denotes beam size. $^\dag$~indicates previously reported scores.}\label{table:llm-present-generative-performance}}
\begin{threeparttable}[width=0.8\textwidth]
    \begin{tabular}{l|rr|rr|rr|rr|rr}
    \toprule
    \multicolumn{1}{c|}{\multirow{2}[2]{*}{\textbf{Model}}} & \multicolumn{2}{c|}{\textbf{Inspec}} & \multicolumn{2}{c|}{\textbf{NUS}} & \multicolumn{2}{c|}{\textbf{Krapivin}} & \multicolumn{2}{c|}{\textbf{SemEval}} & \multicolumn{2}{c}{\textbf{KP20k}} \\
          & \multicolumn{1}{l}{F1@5} & \multicolumn{1}{l|}{F1@M} & \multicolumn{1}{l}{F1@5} & \multicolumn{1}{l|}{F1@M} & \multicolumn{1}{l}{F1@5} & \multicolumn{1}{l|}{F1@M} & \multicolumn{1}{l}{F1@5} & \multicolumn{1}{l|}{F1@M} & \multicolumn{1}{l}{F1@5} & \multicolumn{1}{l}{F1@M} \\
    \midrule
     CatSeq(\#\textit{bs}=1) \citep{yuan-etal-2020-one} & 0.229  & 0.266  & 0.324  & 0.394  & 0.270  & 0.344  & 0.245  & 0.296  & 0.292  & 0.365  \\
     Transformer(\#\textit{bs}=1) \citep{DBLP:conf/acl/YeGL0Z20} & 0.285  & 0.331  & 0.371  & 0.418  & 0.308  & 0.356  & 0.287  & 0.319  & 0.330  & 0.373  \\
       SetTrans(\#\textit{bs}=1) \citep{DBLP:conf/acl/YeGL0Z20} & 0.281  & 0.318  & 0.406  & \textbf{0.452}  & 0.339  & \textbf{0.374} & 0.322  & 0.354 & 0.354  & 0.390 \\
       WR-SetTrans(\#\textit{bs}=1) \citep{DBLP:conf/emnlp/XieWYLXWZS22} & 0.330  & 0.351  & \textbf{0.428} & \textbf{0.452}  & \textbf{0.360}  & 0.362 & \textbf{0.360}  & \textbf{0.370} & \textbf{0.370} & 0.378 \\
     \midrule
    BART-base(\#\textit{bs}=1) \citep{lewis-etal-2020-bart}& 0.270 &	0.323 &	0.366 &	0.424 &	0.270 &	0.336 &	0.271 &	0.321 &	0.322 &	0.388  \\
    BART-large(\#\textit{bs}=1) \citep{lewis-etal-2020-bart}& 0.276 & 0.333 & 0.380 & 0.435 & 0.284 & 0.347 & 0.274 & 0.311 & 0.332 & 0.392  \\
    T5-base(\#\textit{bs}=1)  \citep{DBLP:journals/jmlr/RaffelSRLNMZLL20}&  0.288 & 0.339 & 0.388 & 0.440 & 0.302 & 0.350 & 
 0.295 & 0.326 & 0.336 & 0.388  \\
T5-large(\#\textit{bs}=1)  \citep{DBLP:journals/jmlr/RaffelSRLNMZLL20}& 0.295 & 0.343 & 0.398 & 0.438 & 0.315 & 0.359 & 0.297 & 0.321 & 0.343 & 0.393  \\
KeyBART(\#\textit{bs}=1) \citep{DBLP:journals/corr/abs-2112-08547}& 0.268 & 0.325& 0.373 & 0.430 & 0.287 & 0.365& 0.260 & 0.289 & 0.325 & \textbf{0.398}  \\
    \midrule
zero-shot ChatGPT(\#\textit{bs}=1) & 0.309  & 0.428  & 0.338  & 0.258  & 0.237  & 0.189  & 0.274 & 	0.252  & 0.192 & 	0.158\\ 
1-shot ChatGPT(\#\textit{bs}=1) & 0.421 & 0.480 &	0.355 &	0.359 &	0.297 &	0.298 &	0.319 &	0.326 &	0.298 &	0.295\\
5-shot ChatGPT(\#\textit{bs}=1) & \textbf{0.431} &	\textbf{0.497} &	0.365 &	0.351 &	0.285 &	0.287 &	0.312 &	0.300 & 0.297 &	0.288  \\
\bottomrule
\end{tabular}
\end{threeparttable}
\end{table*}
\begin{table*}[ht]
\centering
\setlength{\tabcolsep}{0.5mm}{
\caption{Results of absent keyphrase prediction using large language models.}
\begin{threeparttable}[width=0.8\textwidth]
    \begin{tabular}{l|rr|rr|rr|rr|rr}
    \toprule
    \multicolumn{1}{c|}{\multirow{2}[1]{*}{\textbf{Model}}} & \multicolumn{2}{c|}{\textbf{Inspec}} & \multicolumn{2}{c|}{\textbf{NUS}} & \multicolumn{2}{c|}{\textbf{Krapivin}} & \multicolumn{2}{c|}{\textbf{SemEval}} & \multicolumn{2}{c}{\textbf{KP20k}} \\
          & \multicolumn{1}{l}{F1@5} & \multicolumn{1}{l|}{F1@M} & \multicolumn{1}{l}{F1@5} & 
          \multicolumn{1}{l|}{F1@M} & \multicolumn{1}{l}{F1@5} & 
          \multicolumn{1}{l|}{F1@M} & \multicolumn{1}{l}{F1@5} & 
          \multicolumn{1}{l|}{F1@M} & \multicolumn{1}{l}{F1@5} & 
          \multicolumn{1}{l}{F1@M} \\
           \midrule
            CatSeq(\#\textit{bs}=1) \citep{yuan-etal-2020-one} & 0.005  & 0.009  & 0.015  & 0.026  & 0.018  & 0.034  & 0.015  & 0.022  & 0.014  & 0.030  \\
              Transformer(\#\textit{bs}=1) \citep{DBLP:conf/acl/YeGL0Z20} & 0.008  & 0.017  & 0.028  & 0.050  & 0.030  & 0.055  & 0.016  & 0.022  & 0.021  & 0.043  \\
           SetTrans(\#\textit{bs}=1) \citep{DBLP:conf/acl/YeGL0Z20}  & 0.018  & 0.029 & 0.041 & 0.061  & 0.046 & 0.073 & 0.029 & 0.035 & 0.035 & 0.056 \\
           WR-SetTrans(\#\textit{bs}=1) \citep{DBLP:conf/emnlp/XieWYLXWZS22} & 0.025  & 0.034  & \textbf{0.057}  & \textbf{0.071}  & \textbf{0.057}  & \textbf{0.074} &\textbf{ 0.040}  & \textbf{0.043 }& \textbf{0.050} & \textbf{0.064 }\\
           \midrule
   BART-base(\#\textit{bs}=1) \citep{lewis-etal-2020-bart}&0.010 &	0.017 &	0.026 	&0.042& 	0.028 &	0.049 	&0.016 &	0.021 &	0.022 &	0.042 \\
BART-large(\#\textit{bs}=1) \citep{lewis-etal-2020-bart}&	0.015 &	0.024 &	0.031 &	0.048 &	0.031 &	0.051 &0.019 	&0.024 &	0.027 &	0.047 \\
T5-base(\#\textit{bs}=1) \citep{DBLP:journals/jmlr/RaffelSRLNMZLL20}&0.011 &	0.020 &	0.027 &0.051 	&0.023 	&0.043 	&0.014 	&0.020 &	0.017 &	0.034  \\
T5-large(\#\textit{bs}=1) \citep{DBLP:journals/jmlr/RaffelSRLNMZLL20}&	0.011 	&0.021 &	0.025 	&0.042 &	0.023 &	0.045 &	0.015 &	0.020 &	0.017 &	0.035 \\
KeyBART(\#\textit{bs}=1) \citep{DBLP:journals/corr/abs-2112-08547}&	0.014 &	0.023 &	0.031 &	0.055 &	0.036 	&0.064 &	0.016 	&0.022 	&0.026 &	0.047  \\
\midrule
zero-shot ChatGPT(\#\textit{bs}=1)& 0.014 &	0.027 &	0.003 &	0.005 &	0.002 	&0.004 & 0.002 & 0.003 & 0.025 & 0.030 \\ 
1-shot ChatGPT(\#\textit{bs}=1)& 0.027 &	\textbf{0.048} &	0.011 &	0.017 &	0.015 &	0.028 &	0.009 &	0.011 &	0.015 &	0.027  \\ 
5-shot ChatGPT(\#\textit{bs}=1)& \textbf{0.028} &	0.046 &	0.010 &	0.015 &	0.016 &	0.031 &	0.016 &	0.021 & 0.015 &	0.027  \\ 
\bottomrule
\end{tabular}
\end{threeparttable}
\label{table:llm-absent-main-performance}
}\end{table*}

\section{Future Directions}
In summary, automatic keyphrase prediction has attracted extensive attention from academia and industry currently. However, it still remains a challenging task in the following aspects:

1) The quality of generated absent keyphrases directly determines the availability of keyphrase generation models. However, dominant models are still unable to produce satisfactory absent keyphrases. Therefore, how to improve the prediction performance on absent keyphrases will be the focus of future research.

2) Intuitively, humans often exploit the information beyond the input document to predict keyphrases. Hence, how to fully exploit more information, such as the extra information from external knowledge base or pre-trained model, for better keyphrase predictions is worth exploring.

3) Short videos have recently emerged as a widespread type of social media due to the explosive growth of the Internet. 
Two new forms of multi-modal information introduced in the search and recommendation scenarios, video and audio, place additional demand on keyphrase prediction. 
Thus, we believe that multi-modal keyphrase prediction is also the future development trend of keyphrase prediction.

4) Existing studies mainly focus on using domain-specific data to train models, such as scientific documents.
However, it is unable to handle different domains of data from the Internet.
Consequently, how to effectively transfer these models to other domains becomes one problem to be solved in practical applications.

5) The conventional evaluation metrics mainly focus on the comparison between the surface representations of stemmed phrases.
However, two phrases may possess the same meaning although their expressions are different.
Hence, the quality evaluation of generated keyphrases should consider the comparison between semantic representations of phrases and the application effect in downstream tasks such as retrieval systems~\citep{DBLP:journals/corr/abs-2106-14726}.

6) Dominant studies model the generations of present and absent keyphrases in a unified manner, although their prediction difficulties vary greatly. Intuitively, it is more reasonable to individually model the generations of absent and present keyphrases. Please note that \cite{wu2022fast} verifies the feasibility of this direction.

7) The generation of keyphrases can draw lesson from the process of human reading and refining keyphrases. 
For example, humans tend to distill the overall idea first and grasp the specifics later,
and thus, an ideal process for keyphrase prediction is to predict keyphrase in a coarse-to-fine manner.

8) Very recently,
ChatGPT has demonstrated effectiveness proficiency across a range of NLP tasks. As such, it is imperative to explore the optimal utilization of ChatGPT in keyphrase prediction, in order to fully exert its remarkable potential.

\bibliographystyle{cas-model2-names}

\bibliography{cas-refs}

\bio{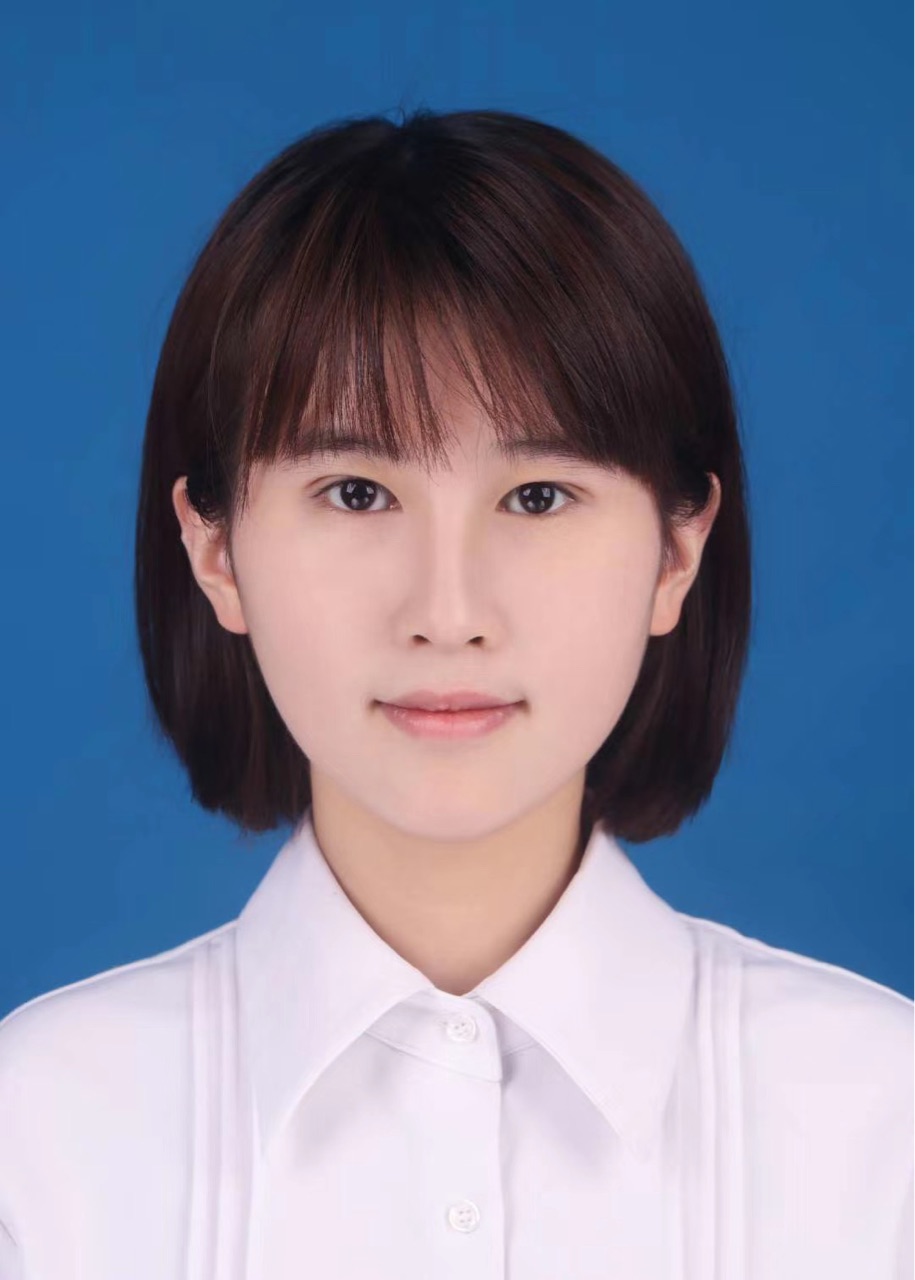}
Binbin Xie received the Bachelor degree in the
school of informatics, Xiamen University, in 2021.
And she is studying for a master’s degree under the
supervision of Prof. Jinsong Su now. Her research
interests include code generation, keyphrase generation and machine translation. \\
\\
\\
\\
\\
\\
\\
\endbio

\bio{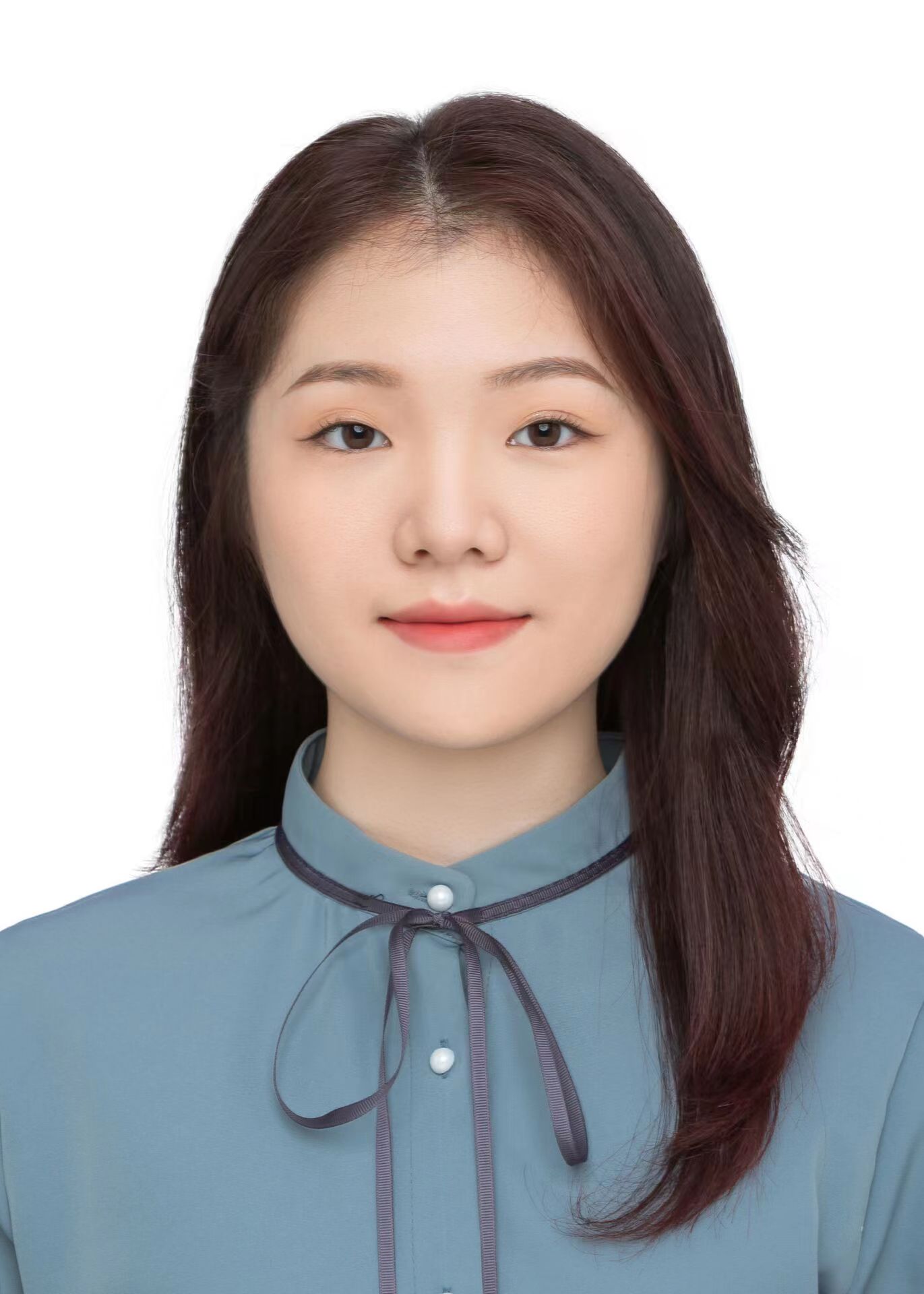}
{Jia Song} was born in 2000. She received her Bachelor degree in Economic Information Engineering School from Southwest University of Finance and Economics, and is a graduate student under the supervision of Prof. Jinsong Su now. Her major research interests are natural language processing and keyphrase generation.
\\
\\
\\
\\
\\
\\
\\
\endbio

\bio{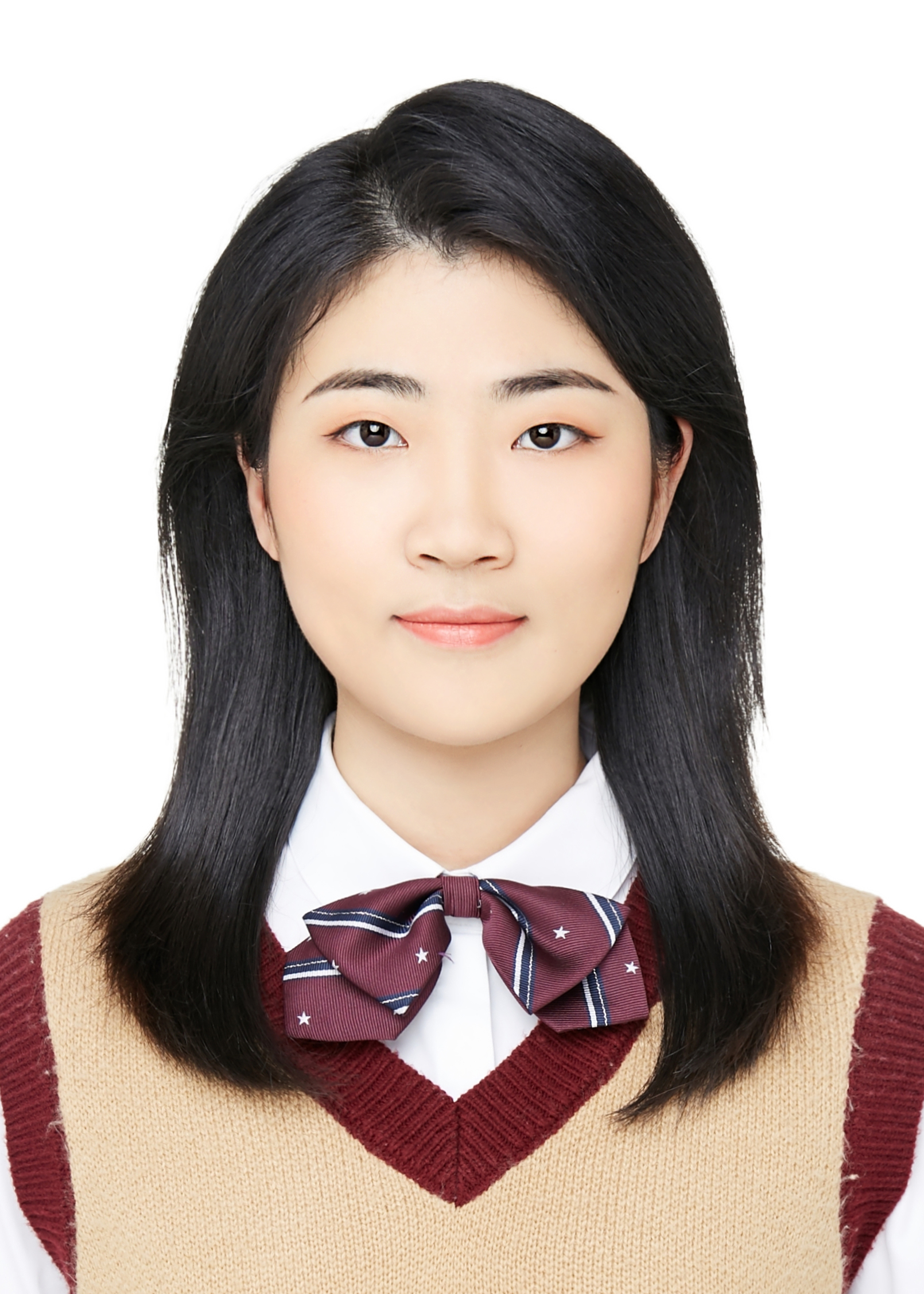}
{Liangying Shao} was born in 2000. She received her Bachelor degree in the school of informatics, Xiamen University, and is a graduate student under the supervision of Prof. Jinsong Su now. Her major research interests are natural language processing and keyphrase generation.
\\
\\
\\
\\
\\
\\
\\
\endbio

\bio{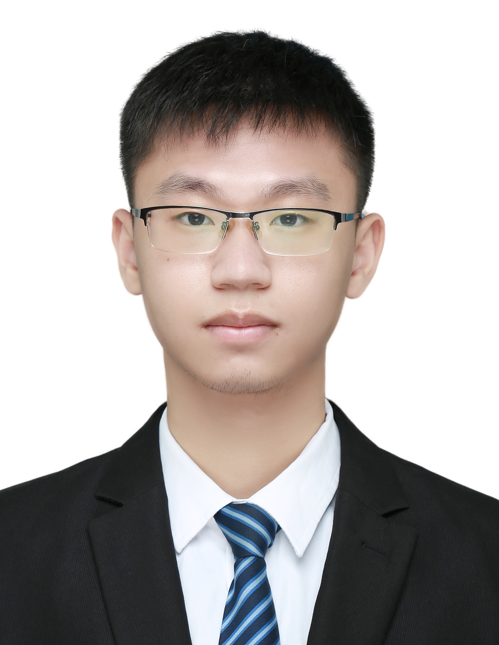}
{Suhang Wu} was born in 2000. He is a undergraduate student at the College of Computer Science and Electronic Engineering of Hunan University now. He will become a graduate student at Xiamen University under the supervision of Prof. Jinsong Su.\\
\\
\\
\\
\\
\\
\\
\\
\endbio

\bio{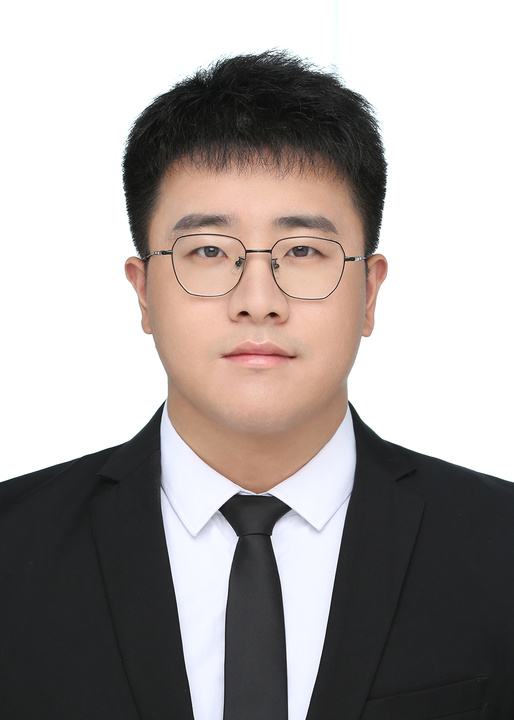}
{Xiangpeng Wei} received the Ph.D. degree from the University of Chinese Academy of Sciences (UCAS) in 2021, supervised by Prof. Yue Hu.
He is now a senior algorithm engineer in the Language Technology Lab at Alibaba DAMO Academy.
His research interests include
natural language processing and neural machine translation.\\
\\
\\
\\
\\
\\
\\
\\
\endbio

\bio{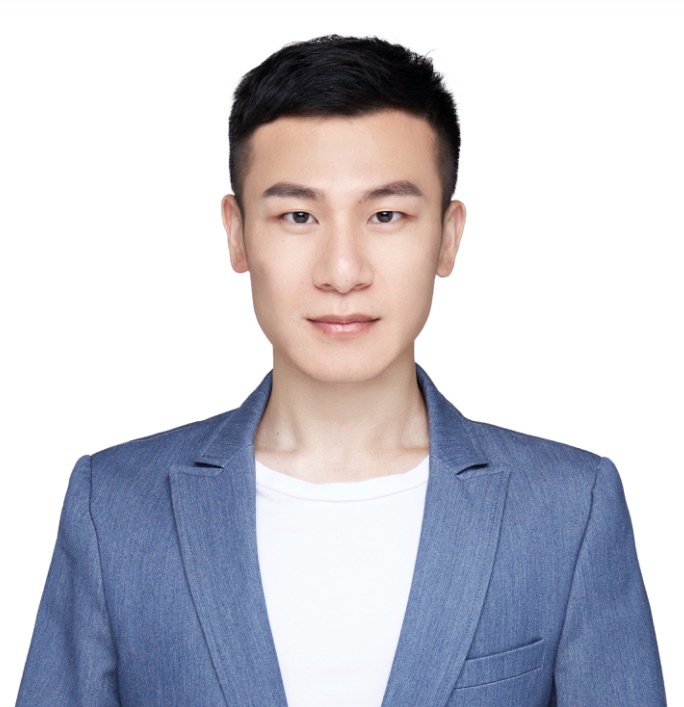}
{Baosong Yang} received the Ph.D at NLP2CT Lab of University of Macau, advised by Prof. Derek F. Wong, and is currently an algorithm expert in the Language Technology Lab at Alibaba DAMO Academy.
His research interests include natural language processing and machine translation. \\
\\
\\
\\
\\
\\
\\
\\
\endbio

\bio{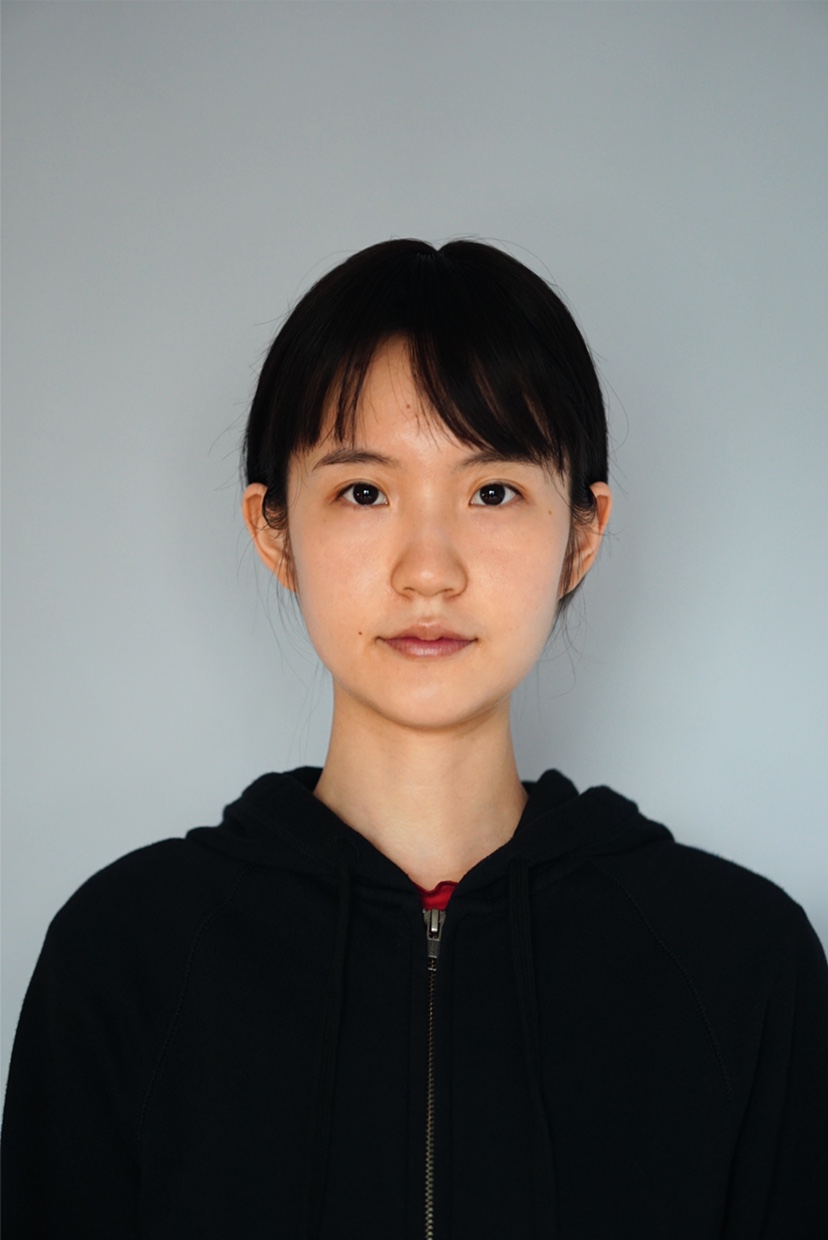}
{Huan Lin} received a master's degree in Xiamen University supervised by Prof. Jinsong Su, and is now an algorithm engineer in the Language Technology Lab at Alibaba DAMO Academy. Her research interests include natural language processing and machine translation.\\
\\
\\
\\
\\
\\
\\
\\
\endbio

\bio{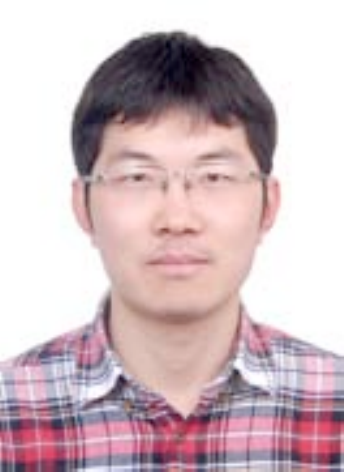}
{Jun Xie} received the Ph.D. degree in computer
science from the Chinese Academy of Sciences,
Beijing, China.
He is currently a senior staff algorithm engineer in the Language Technology Lab at Alibaba DAMO Academy. His research interests include natural language processing and machine translation.\\
\\
\\
\\
\\
\\
\\
\\
\endbio

\bio{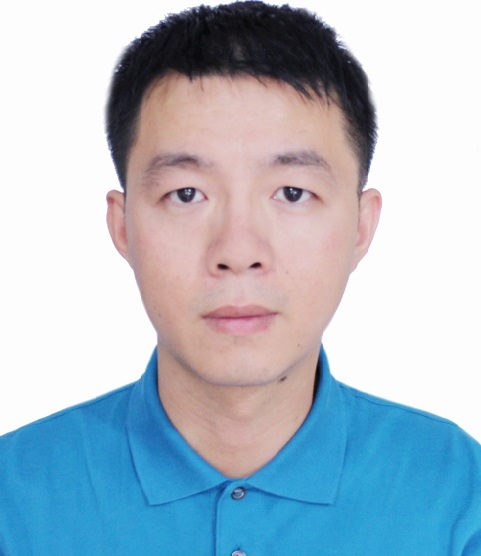}
{Jinsong Su} was born in 1982. He received the Ph.D. degree in Chinese Academy of Sciences, and is now a professor in Xiamen University. His research interests include natural language processing and machine translation.\\
\\
\\
\\
\\
\\
\\
\\
\endbio

\end{document}